\documentclass{article}
\PassOptionsToPackage{numbers,compress}{natbib}
 \usepackage[preprint]{neurips_2026}

\usepackage[utf8]{inputenc} 
\usepackage[T1]{fontenc}    
\usepackage{hyperref}       
\usepackage{url}            
\usepackage{booktabs}       
\usepackage{amsfonts}       
\usepackage{nicefrac}       
\usepackage{microtype}      
\PassOptionsToPackage{table}{xcolor}
\usepackage{tcolorbox}
\usepackage[table]{xcolor}
\usepackage{listings}
\usepackage{amsmath}
\usepackage{wrapfig}
\usepackage{amssymb}
\usepackage{multirow}
\usepackage{fontawesome5}
\lstdefinestyle{pythonstyle}{
    language=Python,
    basicstyle=\ttfamily\footnotesize,
    keywordstyle=\color{blue!50!black}\bfseries,
    commentstyle=\color{gray!80}\itshape,
    stringstyle=\color{teal!60!black},
    showstringspaces=false,
    breaklines=true,
    frame=single,
    framesep=2pt,
    xleftmargin=4pt,
    rulecolor=\color{gray!50},
    backgroundcolor=\color{gray!5},
}
\usepackage{hyperref}
\hypersetup{
    colorlinks=true,
    linkcolor=blue,
    citecolor=blue,
    urlcolor=blue,
}

\lstdefinestyle{logstyle}{
    basicstyle=\ttfamily\footnotesize,
    showstringspaces=false,
    breaklines=true,
    breakatwhitespace=false,
    columns=fullflexible,
    keepspaces=true,
    tabsize=4,
    frame=single,
    framesep=3pt,
    rulecolor=\color{gray!50},
    backgroundcolor=\color{gray!5},
    captionpos=b,
    extendedchars=true,
    inputencoding=utf8,
    escapeinside={(*@}{@*)},
    literate=
        {²}{{$^{2}$}}1  {¹}{{$^{1}$}}1  {°}{{$^{\circ}$}}1
        {₀}{{$_{0}$}}1  {₁}{{$_{1}$}}1  {₂}{{$_{2}$}}1
        {⁻}{{$^{-}$}}1  {⁵}{{$^{5}$}}1
        {·}{{$\cdot$}}1 {×}{{$\times$}}1
        {α}{{$\alpha$}}1 {β}{{$\beta$}}1
        {π}{{$\pi$}}1   {φ}{{$\varphi$}}1
        {—}{{---}}1     {–}{{--}}1
        {‑}{{-}}1       {−}{{$-$}}1
        {→}{{$\to$}}1   {∝}{{$\propto$}}1
        {≈}{{$\approx$}}1
        {’}{{'}}1       {ŷ}{{$\hat{y}$}}1
}

\title{DiscoverPhysics: Benchmarking LLMs for Out-of-the-Box Scientific Thinking}

\author{
  Matt L. Wiemann\thanks{Equal contribution (order by coinflip).}\; \thanks{Work done during internship at NYU/Polymathic AI.}\\
  Princeton University\\
  Polymathic AI \\
  \texttt{ms0821@princeton.edu} \\
  \And
  Lindsay M. Smith\footnotemark[1]\; \footnotemark[2]\\ 
  Princeton University\\
  Polymathic AI \\
  \texttt{lindsay.smith@princeton.edu} \\
  \And
  Peter Melchior \\
  Princeton University \\
  \And
  Siddharth Mishra-Sharma \\
  Boston University \\ 
  \And
  Andrew Gordon Wilson \\
  New York University \\
  \And
  Pavel Izmailov \\
  New York University \\
  \AND
  Carolina Cuesta-Lázaro \\
  Flatiron Institute
  \\
  Institute for Advanced Studies
}

\begin{document}

\maketitle

\begin{center}
\vspace{-0.5cm}
\href{https://github.com/SampsonML/DiscoverPhysics/}{\large\faGithub\ \textcolor{black}{Repository}} \ \ \ \ \ \ \ 
\href{https://huggingface.com/mattWiemann/DiscoverPhysics}{\large\raisebox{-0.3em}{\includegraphics[height=1.35em]{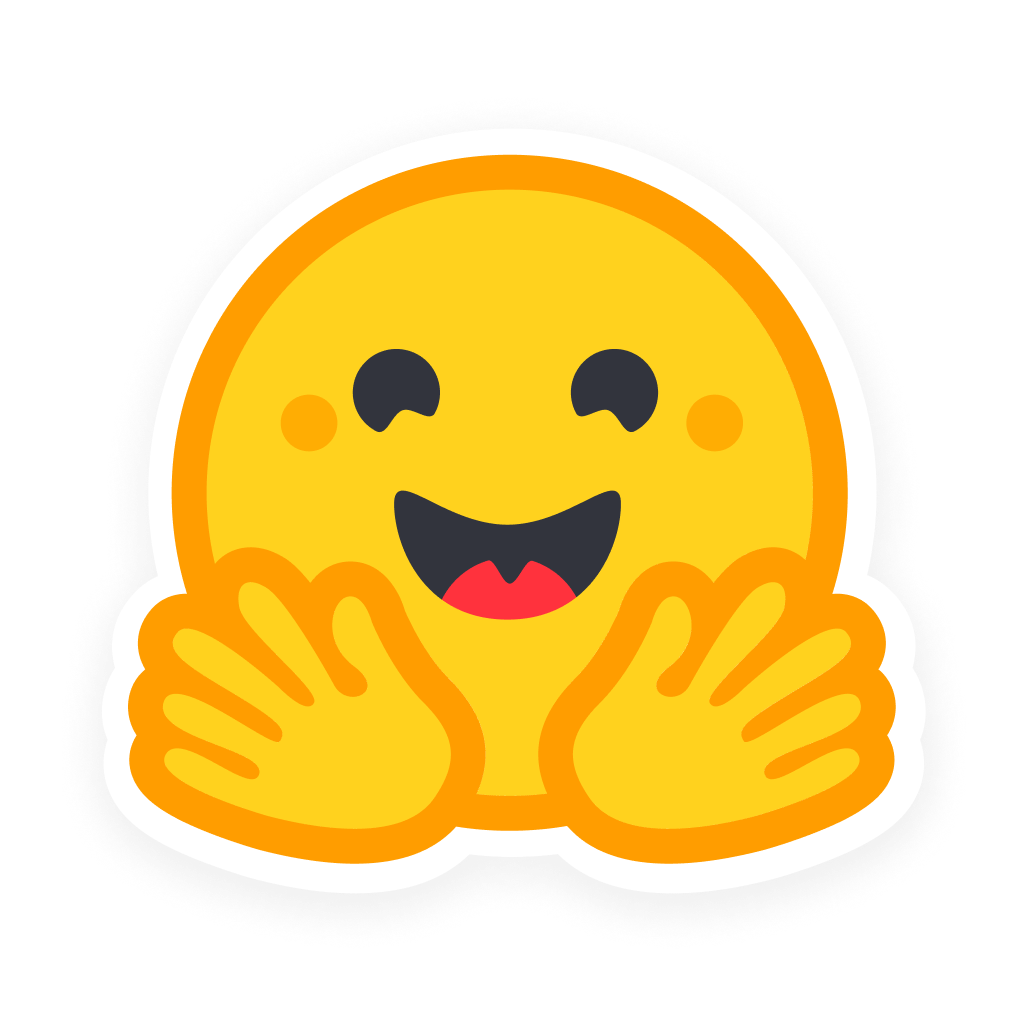}}\ \textcolor{black}{HuggingFace}} \ \ \ \ \ \ \
\href{https://sampsonml.github.io/DiscoverPhysicsLeaderboard/}{\large \faTrophy\ \textcolor{black}{Leaderboard}}
\end{center}
\vspace{0.175cm}

\begin{abstract}
  Frontier LLMs now perform strongly across a wide range of physics evaluations, but it is hard to disentangle genuine reasoning from recall of established science. We introduce \textsc{DiscoverPhysics}, an interactive benchmark that asks a LLM agent to discover the laws of motion of a simulated world whose physics deliberately deviates from our own. We construct 22 worlds governed by, among others, screened and fractional-power gravity, multi-species couplings, hidden dark-matter-like particles, non-coordinate-free physics, and time-varying interactions. Each world is generated on demand by an N-body simulator,
  for which the agent proposes several rounds of experiments, observes raw trajectory data, and ultimately submits both a natural-language explanation of the world's physics and a Python implementation of the inferred law. Because solving a world requires the agent to design informative experiments and revise its hypotheses, the benchmark probes long-horizon reasoning over an experimental history. We evaluate submissions along two complementary axes: trajectory MSE on held-out particles and an LLM-judged explanation score following an expert-written rubric assessing conceptual understanding of each world. Across eleven frontier models, we find that the strongest agents pass only half of the worlds and consistently fail on those where latent structure must be uncovered. Open-source models lag substantially behind commercial models, both in their ability to design informative experiments and in extracting conclusions from the data. We further find that good predictive accuracy does not guarantee high explanation quality and that conceptual understanding depends on hypothesis refinement through well-chosen experiments. 

\end{abstract}

\begin{figure}[t] 
    \centering
    \includegraphics[width=0.97\linewidth]{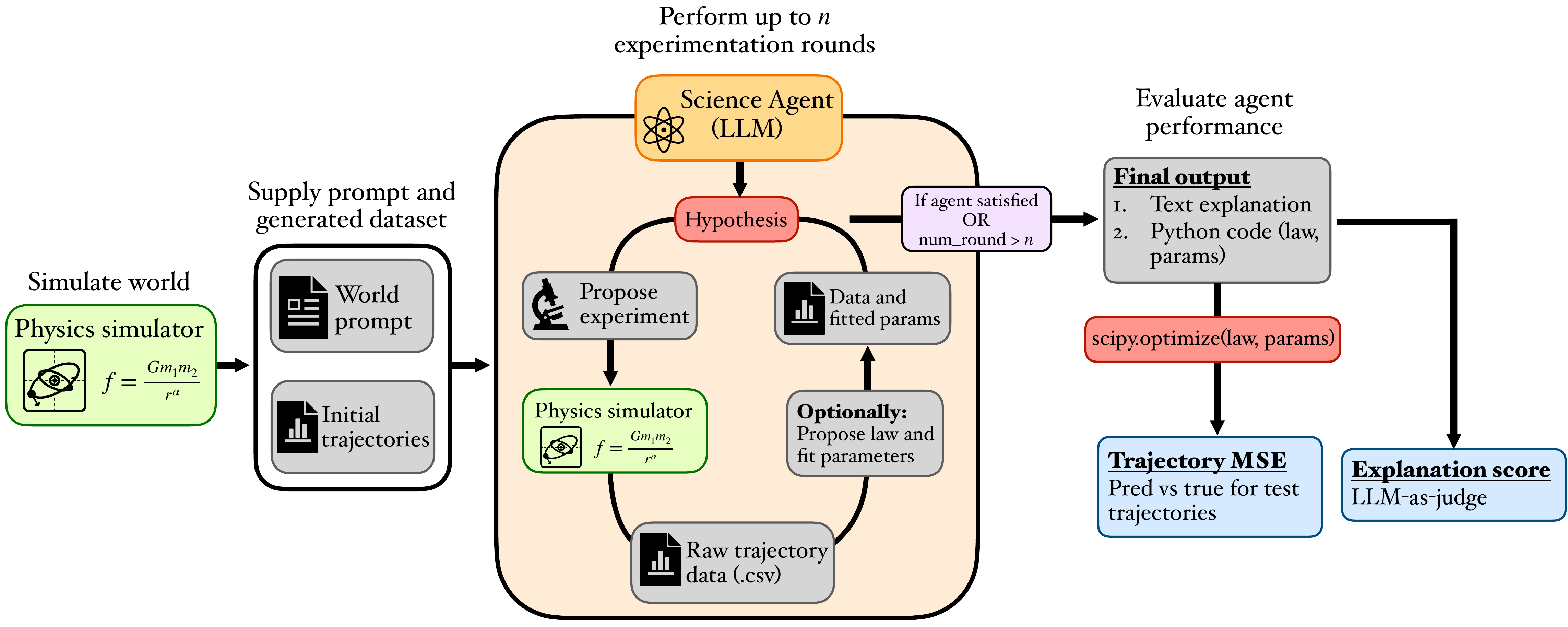}
    \caption{Schematic of the \textsc{DiscoverPhysics} benchmarking pipeline. Our benchmark does not store data products, instead data is generated on demand by a N-body simulator from the equations specified by the world definition. We then generate an initial set of particle trajectories and feed this, as well as the agent prompt about task and experiment instructions, to an LLM. The LLM agent is tasked with discovering the physical laws of the world it is operating in and can propose experiments to be run by the N-body simulator or perform parameter fitting on proposed laws. The agent may experiment for up to $n$ rounds, after which a final law and text explanation must be submitted, both of which are then scored to determine if the agent passed.}
    \label{fig:schematic}
\end{figure}

\section{Introduction}


Large language models (LLMs) already excel at a wide range of specialized tasks like competitive math, coding and graduate-level scientific knowledge. But these represent only a fraction of the skills needed for scientific discovery in the physical sciences, which also requires generating hypotheses, designing experiments, updating beliefs based on noisy observations, and sometimes giving up preconceived notions entirely. Existing evaluations cover slices of the discovery skill space. Knowledge benchmarks measure what a model has learned; reasoning benchmarks measure inference over given inputs; symbolic regression benchmarks measure the recovery of known equation families from fixed datasets. Recent interactive physics benchmarks come closer, but they typically present canonical laws with shifted parameters and a known set of relevant variables, leaving the conceptual part of scientific discovery largely unprobed.


We introduce \textsc{DiscoverPhysics}, an extensible benchmark of simulated worlds whose physics is deliberately non-canonical: short-range exponentially screened forces, fractional Laplacians, hidden particle species, non coordinate-free physics, time-varying couplings, and extra dimensions are some of the examples. Agents interact with each world by placing test particles, observing their trajectories under noise, and iteratively refining their hypothesis about the underlying law. Because the laws governing these worlds are not standard, the agent must identify which features of the observed dynamics are relevant and build mechanistic models that, ideally, predict yet unseen data in this world. The worlds are curated rather than procedurally generated, but they deliberately span a diverse set of non-standard force laws, designed to probe whether LLM agents can discover unusual equations of motion through experimentation and altering their initial guesses.


Our benchmark also seeks to define evaluation metrics for scientific discovery. Predictive accuracy, often the single performance metric for ML tasks, alone does not capture what makes a scientific theory valuable. It also needs to provide a conceptual explanation of a phenomenon that generalizes beyond the observations that motivated it \citep{hogg2024ml4physics}. The history of the natural sciences is full of examples where empirical models fit the data long before the underlying principle was identified. Alchemy attempted to explain the transformation of substances before modern notions of chemistry and elements existed. In physics, Lorentz transformations were derived empirically to explain electromagnetic phenomena before Einstein recognized they follow from a single principle: the invariance of the speed of light. Similarly, energy conservation was long observed in mechanical systems long before Noether's theorem revealed why: conservation laws arise from continuous symmetries. These insights, not just the specific form of the equations, are what allow strong theories to generalize far beyond the initial experiments. \textsc{DiscoverPhysics} therefore evaluates agents along two axes. Trajectory MSE measures how well the agent's discovered law predicts the dynamics on held-out test particles. An explanation score, computed by an LLM judge using a human-written rubric, measures whether the agent has correctly identified the conceptual features of the world that explain those dynamics.


Our contributions are: \begin{itemize}
    \item We introduce \textsc{DiscoverPhysics}, an extensible benchmark of simulated worlds with non-canonical physics that probes scientific discovery through active experimentation, including the design of informative experiments and the iterative refinement of hypotheses.
    \item We define a dual evaluation framework that assesses predictive accuracy (trajectory MSE) and explanatory adequacy (LLM-judged explanation score), capturing whether agents fit the data, identify the underlying principles, or both.
    \item We benchmark 11 frontier and open-source LLMs and find that the strongest agents pass only about half of the worlds, consistently failing on those that require uncovering latent structure such as hidden particle species or extra dimensions.
    \item We show that predictive accuracy and conceptual understanding can decouple: the model achieving the lowest trajectory MSE is not the one with the highest explanation score, and benefits less from additional experimental rounds, which we read as a tendency to fit data without revising its conceptual picture.
    \item We show that open-source models substantially lag frontier models, both in designing informative experiments and in extracting conclusions from the resulting data, with their performance largely unchanged between guided and randomized experimentation.
    \item We release the simulator, public world definitions, and evaluation framework to enable evaluation of future LLMs and community extensions to new worlds.\footnote{For access to full public and private benchmarking suite, request access via HuggingFace: \href{https://huggingface.co/mattWiemann/DiscoverPhysics}{DiscoverPhysics}}
\end{itemize}

\section{Related work}

\textbf{Scientific discovery.} LLMs are employed to accelerate scientific research, from probabilistic frameworks for model discovery \citep{wahl2026probabilistic} to coding agents that evolve algorithms for open but targeted scientific problems \citep{novikov2025alphaevolve}. Systems like The AI Scientist \citep{lu2024aiscientist} take a most ambitious framing, automating the full research lifecycle including ideation, experimentation, and writeup. These works aim to establish that LLMs can meaningfully participate in scientific workflows, but their evaluation typically targets paper-level quality judged by LLMs rather than recovery of a ground-truth mechanism, leaving open the question of how well models perform discovery against an objective standard.

\textbf{Knowledge benchmarks.} Static benchmarks of scientific knowledge have served as a primary measure of LLM capabilities in technical domains. GPQA \citep{rein2023gpqa} provides 448 expert-validated, "Google-proof" graduate-level questions in physics, chemistry, and biology, with a rigorous validation procedure in which domain PhDs reach 65 -- 74\% accuracy. Current state-of-the-art LLMs achieve over 90\% accuracy. However, strong performance on GPQA reflects what a model has learned about known science, but not what it can discover.

\textbf{Reasoning benchmarks.} A complementary line of work targets reasoning rather than memorized knowledge. ARC \citep{chollet2019measure} introduces few-shot puzzles explicitly designed to resist memorization through novel task generation. While humans solve ARC tasks with near-perfect accuracy, AI systems show wide performance variation, making it one of the most demanding tests of general reasoning. Whilst ARC measures reasoning, scientific discovery requires reasoning from experimentation in noisy environments, where the relevant features must themselves be inferred.

\textbf{Symbolic regression and equation discovery.} A long line of work evaluates the recovery of physical equations from data. Recently, LLM-based approaches such as LLM-SR \citep{shojaee2024llmsr} use language models as priors to accelerate equation discovery, and LLM-SRBench \citep{shojaee2025llmsrbench} evaluates these capabilities while avoiding memorization of known physics. More recent agentic frameworks such as KeplerAgent \citep{yang2026think} infer physical structure such as symmetries from data and translate it into constraints, restricting the search space of downstream symbolic regression algorithms. Our goals depart from symbolic regression more fundamentally: scientific discovery is not merely about retrieving an equation that fits the data. Symbolic regression methods often interpolate observations through complex functional forms, which bear little resemblance to the generative mechanism. Even when a concise equation is recovered, the act of discovery then depends a human scientists to recognize the conceptual content of the underlying physics. Our benchmark evaluates conceptual understanding in addition to trajectory MSE, asking not only whether the agent produces a law that predicts the dynamics but whether it has identified the structural features that explain them.

\textbf{Interactive scientific discovery.} A growing body of work evaluates the  scientific capabilities of LLMs through interactive worlds. DiscoveryWorld \citep{jansen2024discoveryworld} introduces 120 scientific tasks in a text-based simulator and proposes an evaluation scheme that separates task completion from explanatory knowledge graded against gold references. Other benchmarks evaluate data-driven hypothesis search \citep{majumder2024discoverybench}, agent performance on tasks from peer-reviewed publications \citep{chen2025scienceagentbench}, closed-loop experimentation in biology \citep{roohani2024biodiscoveryagent, mitchener2025bixbench}, and causal-graph discovery via interventions \citep{chen2025autobench}. \citet{riosgarcia2026aiscientists} further analyze agent reasoning traces across eight scientific domains and find that evidence is ignored in 68\% of traces and refutation-driven belief revision occurs in only 26\%, even when agents arrive at correct answers. This suggests that outcome-based metrics miss systematic reasoning failures. While these works establish that interactive evaluation is essential, none directly tests whether an agent recovers a ground-truth generative mechanism in a setting where the underlying physics is genuinely unusual.

\textbf{Physics discovery benchmarks.} The closest precedents to our work specifically target physics. NewtonBench \citep{zheng2025newtonbench} introduce 324 tasks across 12 physics domains via alterations of canonical laws. They report that frontier models reach roughly 30\% accuracy on complex tasks, with sperformance halving under minimal observational noise. PhysGym \citep{chen2025physgym} provides 97 interactive physics problems with controlled prior-knowledge levels, allowing the contribution of a model's prior physical knowledge to be cleanly separated from its discovery ability. Gravity-Bench \citep{koblischke2025gravity}, focuses on agentic discovery in two-body gravitational dynamics with budget-limited observations. In all these benchmarks, the structure of the discovery problem is largely given. NewtonBench shifts the parameters of canonical laws while preserving their functional family, PhysGym exposes named controllable variables and asks for closed-form relationships in those variables, and Gravity-Bench restricts to two-body dynamics where the relevant degrees of freedom are evident. Real scientific breakthroughs often rely on identifying which variables matter, recognizing hidden degrees of freedom, and distinguishing signal from noise. Our worlds are designed to probe these capabilities.



\section{The DiscoverPhysics Benchmark}
This benchmark suite evaluates whether a LLM agent can infer the physical laws governing a simulated world through active experimentation. Because all trajectories are generated at run-time by a simulator, the benchmark depends on no static data products and can be extended to arbitrary force laws, noise levels, and particle counts. Each evaluation pairs an agent with a world whose force law is hidden, an experimental interface for proposing and running experiments, and two final metrics: predictive accuracy on held-out trajectories and explanatory agreement with a human-written rubric.

\subsection{Overview}

The unit of evaluation is a single session in which the agent is given a fixed budget of experimentation rounds to discover the hidden law of one world. Within this budget, the agent can freely alternate between running experiments and proposing candidate laws. At the end of the budget it must commit to a final answer.

A single experimental round works as follows. The agent submits a list of test particles, each specified by an initial position, initial velocity, charge, and a list of measurement times. The simulator integrates these particles under the world's hidden law together with any pre-existing particles in the world, and returns the requested positions and velocities at each measurement time. The agent may also propose a candidate law and request a parameter fit against the trajectories observed so far. 

Once the round budget is exhausted, the agent submits two final outputs: a short natural-language explanation of the world's physics and a Python implementation of the inferred law. These are then scored on trajectory MSE against held-out test particles and on explanation quality judged against a human-written rubric (see \autoref{sec:eval_metrics}).



The interactive experimentation loop (visualized in \autoref{fig:schematic}) is central to the benchmark because the problems are not solvable as pure curve fitting. The agent must decide which regions of phase space to probe, which quantities to vary, and whether to intervene by changing charges, placing probes at short or long range, or repeating experiments at different evaluation times. 

To isolate experimental design from reasoning over existing data, we provide an optional \textit{randomized} mode in which the per-round particle initial conditions are sampled from a uniform grid rather than chosen by the agent. The agent still observes the resulting trajectories and proposes laws as usual, so any performance gap between the two modes can be attributed to the agent's choice of experiments.


Every discovery session opens with the same universal prompt:

\begin{tcolorbox}[
    colback=gray!15!white,
    colframe=gray!95!white,
    sharp corners,
    boxrule=1pt,
    title=Universal Prompt,              
    fonttitle=\bfseries,          
    left=4pt, right=4pt,        
    top=4pt, bottom=4pt,      
]
\textit{``You are an expert physicist and AI research scientist tasked with discovering scientific laws in a simulated universe. Your goal is to propose experiments, analyze the data they return, and ultimately deduce the underlying scientific law. Please note that the laws of physics in this universe may differ from those in our own. You can perform experiments to gather data, but you must follow the protocol strictly.''} 
\end{tcolorbox}

Thereafter we provide a short, world-specific protocol describing the JSON schema for experiments and how to invoke the parameter-fitting tool.
Crucially, the agent receives no information about the world's force law, field equation, source coupling, or the existence of hidden particle types. All of these must be inferred from the trajectories returned by the simulator.


\textbf{Final output.} For evaluation, the models must give a final output that includes both a short plain-text explanation of the physical laws that govern the world and a physical law in the form of a Python function definition to evolve particles according to the discovered law. Along with the function definition, the agent may present up to five parameters, which will then be fit to the observable data before the final code is executed. We show an example output below, where we exclude the majority of the python code for brevity. Full example outputs for each world are in \autoref{sec:appendix_examples}.

\begin{lstlisting}[style=pythonstyle, basicstyle=\ttfamily\small, frame=single, framesep=2pt, xleftmargin=4pt]
def discovered_law(pos1, pos2, p1, p2, velocity2, duration, **params):
    """Particle 2 is accelerated toward particle 1 by a radial force per unit along r^ that combines a static source term G*p1, a radial-velocity term (analogous to an advective/retardation coupling), and a centripetal-like tangential kinetic-energy term, all scaled by 1/r and independent of p2."""
    ...
    return trajectories

def fit_parameters():
    return {
        "G":     {"init": 1.0, "bounds": [0.1, 5.0]},
        ...
    }
\end{lstlisting}

\textbf{Agentic experimentation and tool use.}
Importantly, we consider the agents to be an active participant in these worlds. As such, we provide the ability to run experiments, which are critical to determine the laws of the world. The experiments are conducted via a N-body simulator we provide. 
Each experiment is defined by JSON-formatted text, which specifies initial conditions for the controllable particles plus a list of measurement times. The simulator integrates from $t=0$ (or an optional \texttt{start\_time}) and returns positions and velocities at each requested time. Trajectories from every experiment are appended to a per-run CSV log. The agent can use an optional  tool that performs a least-squares fit of the parameters of its current candidate law against this CSV log using \texttt{scipy.optimize}. We use the same tool to evaluate the final MSE. The fit results are returned in the form of a text prompt which is given to the science agent at the beginning of the next round.


\subsection{Physics simulator and world construction }
The worlds are defined by their pairwise force laws, evolved by an N-body simulator with interchangeable integrators: RK4, 4th- and 6th-order Yoshida \citep{yoshida1990construction}, Dormand-Prince (\texttt{jax.experimental.ode.odeint}) \citep{jax2018github}, symplectic Euler (kick-drift), and leapfrog (velocity-Verlet). This lets the user (but not the agent) trade simulation precision against compute. We use Yoshida4 by default because it is symplectic, 4th-order accurate, and fast under JIT compilation. A small softening term is added to avoid numerical instabilities.


Each particle carries a kinematic state: position $\mathbf{r}_i$ and velocity $\mathbf{v}_i$, together with a set of generalized charges $\mathbf{q}_i$ that determine how it couples to forces. (In physics, a charge is any intrinsic property that sets a particle's response to a given interaction: e.g. mass is the charge of the gravitational force). The pairwise force takes the general form

\[
\mathbf{F}_{i \leftarrow j}
\;=\;
F_{\mathrm{mag}}\!\left(r_{ij},\,\mathbf{q}_i,\,\mathbf{q}_j,\,t\right)\,\hat{\mathbf{r}}_{ij},
\]

where $r_{ij} = \|\mathbf{r}_i - \mathbf{r}_j\|$ and $\hat{\mathbf{r}}_{ij}$ is the unit separation vector. 

Letting $\mathbf{q}_i$ be a vector rather than a scalar lets a single world contain multiple particle species with distinct couplings. For instance, three species with distinct pairwise interactions. The explicit time dependence accommodates time-modulated couplings, as in our oscillator world, where the coupling strength periodically reverses sign. In our implementation we decompose $\mathbf{q}_i \equiv (s_i, c_i)$ into a source charge $s_i$ controlling how strongly particle $i$ generates the field, and a response charge $c_i$ controlling how strongly it feels the field generated by others. This lets a single world contain pure sources ($c_i = 0$), neutral test probes ($s_i = 0$), and ordinary particles that both source and respond ($s_i = c_i$) without separate code paths. When $s_i = c_i$ for all particles, the formalism reduces to a standard symmetric pairwise interaction.

\begin{figure}[t]
    \centering
    \includegraphics[width=0.98\linewidth]{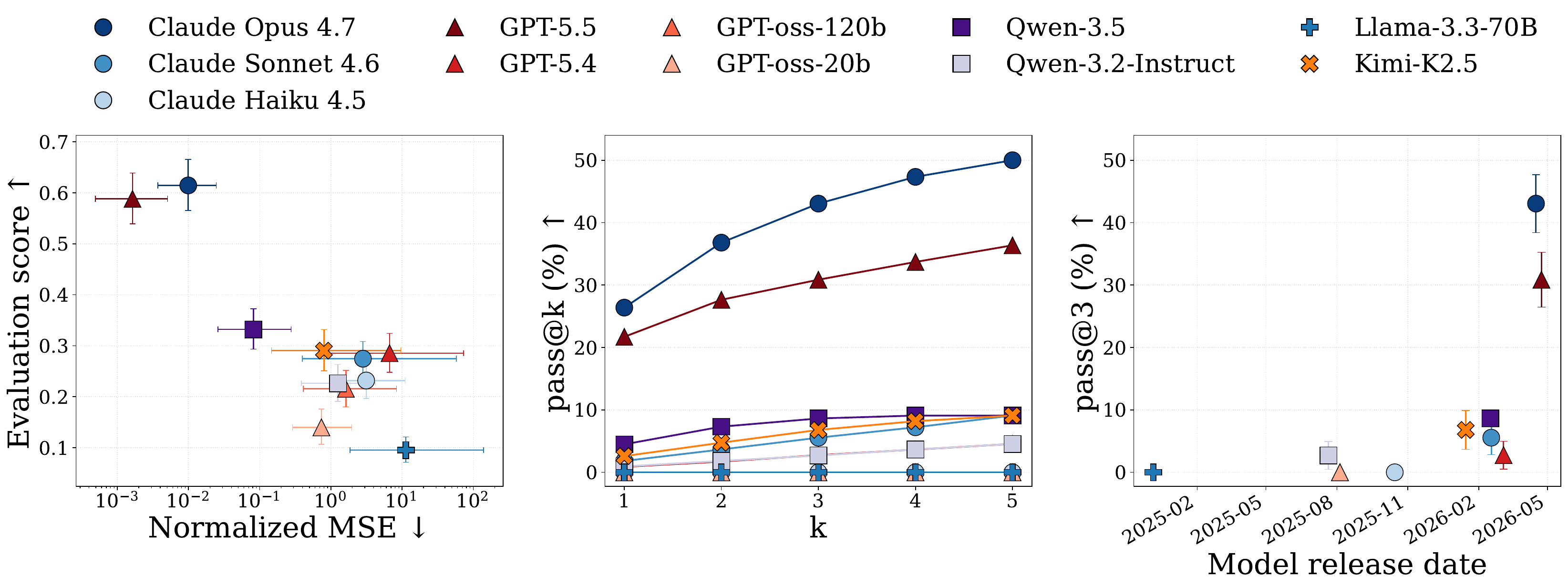}
    \caption{Performance metrics of the suite of 11 tested LLM models. From left to right, we show a Pareto plot of evaluation score as a function of normalized MSE, expected pass@k as a function of $k$ attempts, and expected pass@3 as a function of model release date.}
    \label{fig:money}
    \vspace{-1em}
\end{figure}

Because the simulator evaluates pairwise interactions directly, its computational cost scales as $\mathcal{O}(N^2)$. 
For the modest particle counts used in our worlds (typically $N \lesssim 35$), the computational cost is low. The benefit of this design is flexibility: arbitrary, non-standard interaction laws can be specified without solving a corresponding PDE. Optionally, Gaussian noise can be added to simulator outputs, controlled by a CLI parameter; the noise level used in our main results (\autoref{tab:main_table}, \autoref{fig:money}) is $\sigma = 0.05$, representing $5\%$ observation noise compared to the test particle variance, while held-out test trajectories used at evaluation time are noise-free.



A full description of the worlds' governing principles and parameters is given in \autoref{sec:appdescr}.

\subsection{Evaluation metrics}
\label{sec:eval_metrics}

\textbf{Trajectory MSE.} Once the agent submits its final law, we fit its free parameters to the data the agent has already observed and roll out the resulting law on a set of held-out test particles. The trajectory MSE is the mean squared Euclidean distance between true and predicted positions across these held-out trajectories. The held-out trajectories are noise-free. We report the geometric mean across worlds to reduce the influence of occasional large outliers.

\textbf{Explanation score.} We present each physics world with a rubric of human-labeled explanation scores from $0 \to 10$. An example of a rubric is given in \autoref{sec:apprubric}. After the science-agent presents its final text summary of the physical laws of the world we use a separate LLM (always \texttt{claude-opus-4-6}, which is not in our reported baselines) to compare the given explanation to the human written rubric and provide a score. The explanation score is a proxy for the conceptual understanding of the system.

\textbf{Worlds pass@$k$.}
We report the expected worlds passed from $k$ independent attempts as pass@$k$. A world is considered a pass if the per-trajectory normalized MSE is below 0.1 ($10\%$ error) and the explanation score is $\geq 0.9$. The MSE normalization is done by calculating the variance of the test particle trajectories and dividing the MSE accordingly to account for the difference in total particle travel in different worlds. To calculate pass@$k$ we randomly sample without replacement from each of the 5 attempts per model,  $k$ times per world, and a single passing attempt per world is sufficient to count the world as passed. For instance, pass@3 is computed by randomly sampling 3 seeds per world, recording how many worlds had at least one pass. We repeat the sampling 1,000 times to calculate the expectation value and standard errors, which are reported in \autoref{tab:main_table}. 

\textbf{Held out worlds.}
We publicly release the setup and evaluation rubrics for 11 public worlds. However, to ensure the validity of this benchmark, we keep the remaining 11 worlds private. All results are shows over the full suite of 22 worlds, so we redact the names of the private worlds in all figures and results.


\begin{table}[t] 
    \centering
    \renewcommand{\arraystretch}{1.3}
    \setlength{\tabcolsep}{3pt}
    \caption{Results of the baseline testing of 11 LLM models on the \textsc{DiscoverPhysics} benchmark, reporting the mean explanation score, per-particle mean position error, and the expected percentage of worlds pass@$k$ as a percentage. }
    \footnotesize
    \begin{tabular}{l@{\hskip 6pt}ll@{\hskip 6pt}lllll}
        \toprule
        Model & $\langle\text{Explanation}\rangle$ & norm $\langle\text{MSE}\rangle$ & $\rm{pass}@1 $ & $\rm{pass}@2$ & $\rm{pass}@3$ & $\rm{pass}@4$ & $\rm{pass}@5$\\
        \midrule
        \multicolumn{8}{l}{\textit{\textcolor{blue!70!black}{Proprietary}}} \\
        claude-opus-4-7   & $\mathbf{0.61\pm.03}$ & $0.01^{.006}_{.004}$    & $\mathbf{26.4\pm.3}$ & $\mathbf{36.8\pm.2}$ & $\mathbf{43.1\pm.2}$ & $\mathbf{47.2\pm.1}$  & $\mathbf{50.0}$ \\
        claude-sonnet-4-6 & $0.28\pm.02$ & $2.8^{10.4}_{2.6}$  & $1.8\pm.1$ & $3.7\pm.1$ & $5.5\pm.1$ & $7.2\pm.05$& $9.1$ \\
        claude-haiku-4-5  & $0.23\pm.02$ & $3.1^{2.6}_{1.4}$  & $0.0$ & $0.0$ & $0.0$ & $0.0$  & $0.0$ \\
        gpt-5.5           & $0.59\pm.03$ & $\mathbf{0.002}^{.001}_{.001}$ &$21.7\pm.2$ & $27.6\pm.2$ & $30.8\pm.1$ & $33.7\pm.1$ & $36.4$  \\
        gpt-5.4           & $0.29\pm.03$ & $6.7^{13.9}_{4.5}$  & $0.9\pm.1$ & $1.7\pm.1$ & $2.7\pm.1$ & $3.6\pm.1$  & $4.5$ \\
        \midrule
        \multicolumn{8}{l}{\textit{\textcolor{green!50!black}{Open-source}}}\\
        gpt-oss-120b      & $0.22\pm.02$ & $1.6^{1.9}_{0.9}$    & $0.0$ & $0.0$ & $0.0$ & $0.0$  & $0.0$ \\
        gpt-oss-20b       & $0.14\pm.02$ & $0.7^{0.5}_{0.3}$  & $0.0$          & $0.0$          & $0.0$          & $0.0$         & $0.0$ \\
        Qwen3-235B-A22B   & $0.24\pm.02$ & $1.0^{0.9}_{0.5}$ & $1.0\pm.1$ & $1.8\pm.1$ & $2.7\pm.1$ & $3.7\pm.1$ & $4.5$ \\
        Qwen3.5-397B-A17B & $0.33\pm.03$ & $0.1^{.07}_{.04}$  & $4.5\pm.1$ & $7.3\pm.1$ & $8.6\pm.01$ & $9.1$ & $9.1$ \\
        Kimi-K2.5 & $0.29\pm.02$ & $0.8^{1.1}_{.4}$  & $2.6\pm.1$ & $4.8\pm.1$ & $6.8\pm.1$ & $8.2\pm.1$ & $9.1$ \\
        Llama-3.3-70B & $0.10\pm.01$ & $14.1^{35.8}_{10.1}$  & $0.0$ & $0.0$ & $0.0$ & $0.0$ & $0.0$ \\
        \bottomrule
    \end{tabular}
    \label{tab:main_table}
\end{table}

\section{Experiments}
We present the results of benchmarking eleven pretrained LLMs acting as science agents, using the API-default reasoning settings for \texttt{gpt-5.5} (medium) and \texttt{claude-opus-4-7} (high), which are roughly at the mid-point of the reasoning effort spectrum; higher reasoning effort could potentially yield higher scores on both models, and the numbers reported here should be read as a lower bound at default settings rather than a ceiling. API-default sampling temperature is used as well. For each of the 22 physical worlds, we gather results across five seeds, presenting the mean and bootstrapped standard error for all reported results with 5,000 bootstrap resamples. 

\textbf{Model performance comparison.} In \autoref{tab:main_table} we show the full results for our benchmark over a series of open and proprietary models for the mean explanation score, normalized MSE, and the expected worlds passed with $k$ attempts. This data is visualized in \autoref{fig:money}. \texttt{claude-opus-4-7} performs best in terms of explanation score for each world and the expected pass rate at all $k$ attempts, while \texttt{gpt-5.5} has the overall lowest MSE.

\begin{figure}
    \centering
    \includegraphics[width=0.99\linewidth]{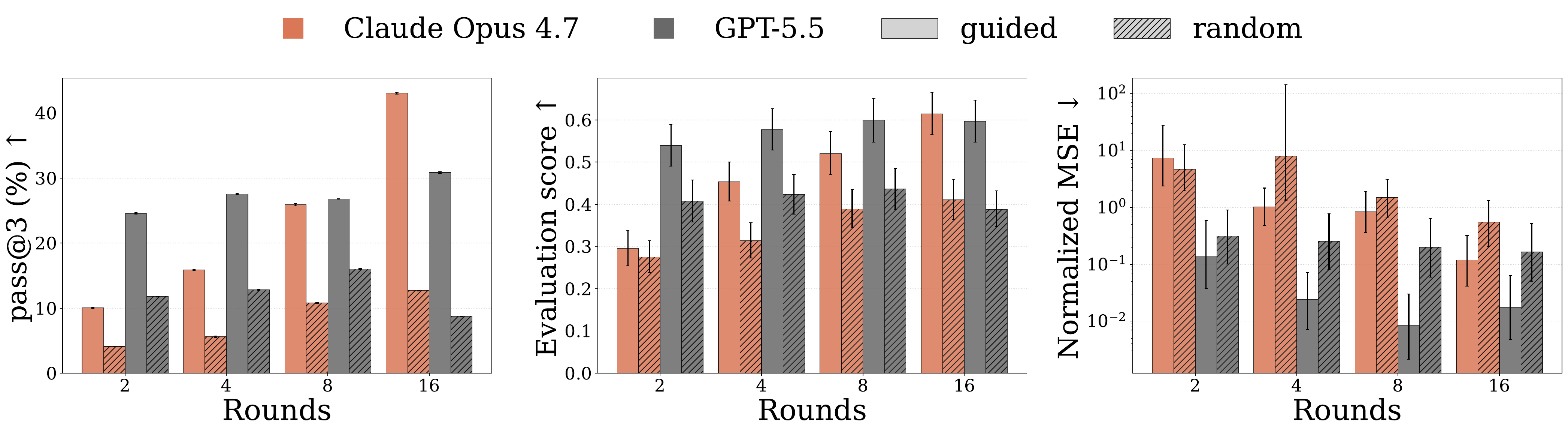}
    \caption{Performance metrics of \texttt{claude-opus-4-7} and \texttt{gpt-5.5} as a function of allowed experimental rounds. We show (from left to right) the expected pass rate at $k=3$, explanation score averaged over all worlds, and normalized MSE averaged over all worlds. We also show the results of random experimentation in the cross-hatched colorbars for both models.}
    \label{fig:guided_vs_rng} 
    \vspace{-0.5cm}
\end{figure}

\textbf{Guided vs random experimentation.} In \autoref{fig:guided_vs_rng} we compare the performance of \texttt{gpt-5.5} and \texttt{claude-opus-4-7} as a function of experimental rounds. We see qualitative differences between the two models. For the guided experiments, \texttt{claude-opus-4-7} shows monotonic improvement across all 3 performance metrics as the number of experimental rounds increases, with the final pass rate at $k=3$, and evaluation score higher than \texttt{gpt-5.5}, however with a higher mean MSE. In contrast, \texttt{gpt-5.5} initially far exceeds the performance of \texttt{claude-opus-4-7} at the lower round budgets of 2, 4, and 8 in both explanation score and MSE, but sees no significant improvement in the pass rate or explanation score after round 4, while MSE still continue to decrease. Notably, at round budgets of 2, 4, and 8, the judge rates \texttt{gpt-5.5}'s explanations strictly higher than \texttt{claude-opus-4-7}s, with \texttt{opus-4-7} only surpassing \texttt{gpt-5.5} at 16 rounds on the explanation axis. A judge biased toward its own model family would be unlikely to produce this pattern across three of four round budgets. 

We also compare agent-guided and randomly selected experiments at each round. In the LLM-guided experiments, the agent chooses the positions of the particles it tests in each experimental round. In the random experiments, the positions are chosen at random. The agent still sees the results of the time evolution of the particles from the simulator. This ablation determines whether the agents are capable of choosing insightful experiments to aid their analytic capabilities.
\autoref{fig:guided_vs_rng} indicates that \texttt{claude-opus-4-7} uses its experiments more wisely and learns more from new data than \texttt{gpt-5.5}.
We note due to the unavailability of the exact amount of reasoning/thinking tokens used by the OpenAI models, we did not perform an analysis of performance as a function of tokens.

\begin{figure}[t]
    \centering
    \includegraphics[width=0.98\linewidth,trim={0 0.5cm 0 0},clip]{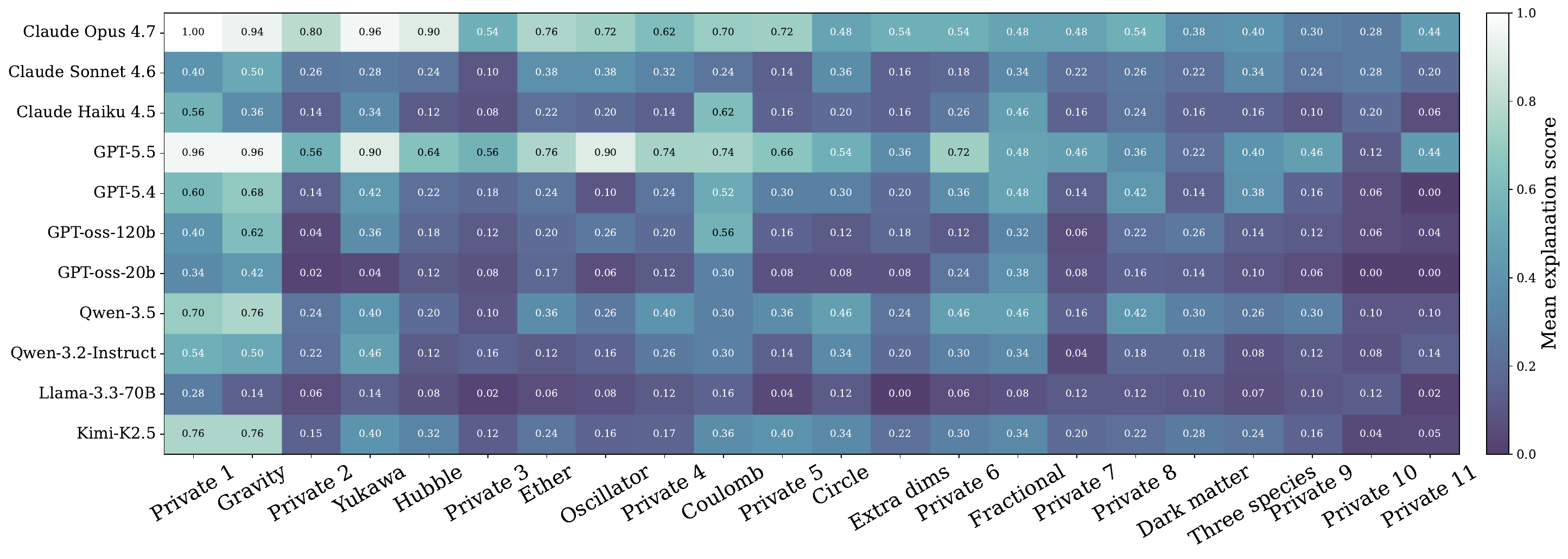}
    \includegraphics[width=0.98\linewidth]{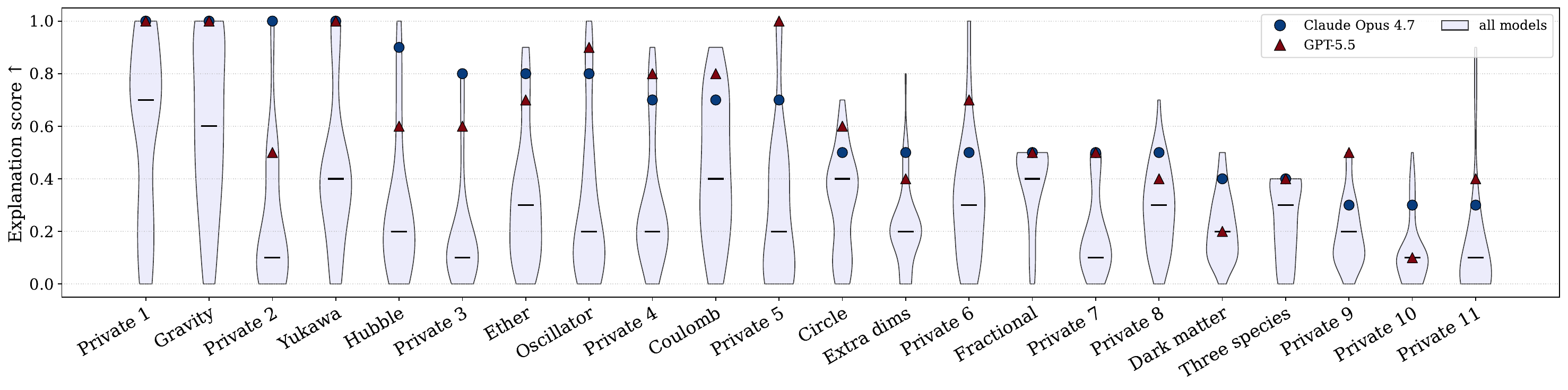}
    \caption{\emph{Top:} Mean explanation score per model (rows) and world (columns).
    The cell values show the mean explanation score between $0\to1$, averaged over all 5 seeds for each model.
    \emph{Bottom:} Distribution of explanation scores by world, pooled across all models and seeds. Each violin shows the score distribution for one world; the horizontal bar marks the median. Worlds are sorted left to right by from easiest to hardest based on \texttt{claude-opus-4-7} performance.}
    \label{fig:world_heat}
\end{figure}

\textbf{World difficulty.} \autoref{fig:world_heat} shows a heatmap of the mean explanation score across five attempts for every physics world. The worlds are ordered from easiest to hardest based on median explanation scores for the strongest model. While the oscillator world proved most difficult for most models, \texttt{gpt-5.5} and \texttt{claude-opus-4-7} perform well on this test, while they struggle more than others on the \texttt{coulomb} world, demonstrating that their explanatory power is not uniformly higher.

\begin{figure}[b]
    \vspace{-0.5cm}
    \centering
    \includegraphics[width=0.99\linewidth]{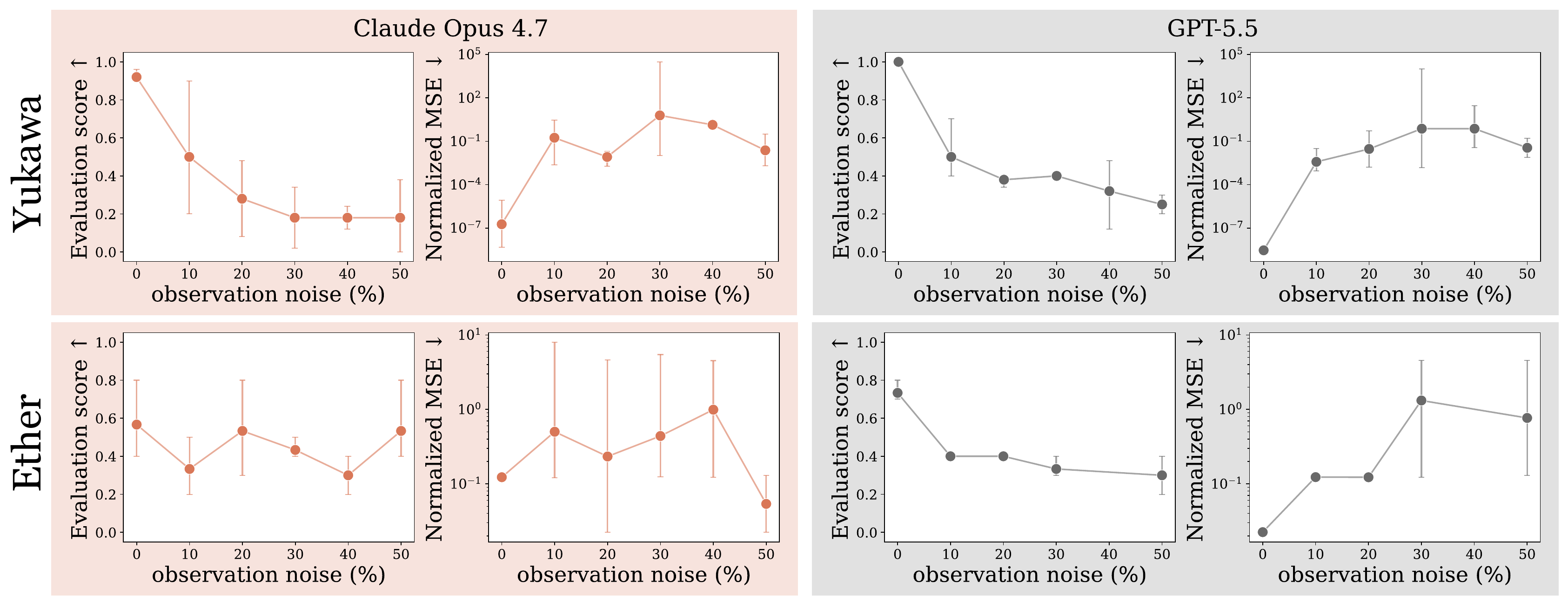}
    \caption{Noise ablations over the Yukawa (top) and Ether (bottom) worlds for \texttt{claude-opus-4-7} and \texttt{gpt-5.5}. We show the evaluation score, and normalised MSE as a function of observation noise added as a fraction of total trajectory variance (as in the MSE normalization scaling).}
    \label{fig:noise-ablation} 
    \vspace{-1em}
\end{figure}

\textbf{Noise ablation.} \autoref{fig:noise-ablation} shows a noise ablation on two worlds (Yukawa and Ether) for our top models. Increasing observational noise generally reduces the explanation score and increases MSE, but with notable per-world differences. Yukawa contains only 2 particles versus Ether's 26, so Gaussian noise degrades the Yukawa signal much faster, the explanation scores collapse beyond $20\%$ noise. Surprisingly, \texttt{claude-opus-4-7} on Ether retains high explanation scores even at $50\%$ noise. In the output logs we see for these trials the model recognises the noise and designs experiments that mitigate its effect; we include a sample response in \autoref{sec:appendix-noise}. Note that this ablation adds Gaussian noise to both position and velocity, whereas the main results in \autoref{tab:main_table} and \autoref{fig:money} use position-only noise; we chose the harsher regime to stress-test the ablation, and expect model rankings to be preserved under either convention.

\section{Discussion}

The benchmark presented here is designed to reveal the progress and the limitations of current LLMs to perform open-ended discovery of physical laws through experimentation. The strongest models we tested, \texttt{claude-opus-4-7} and \texttt{gpt-5.5}, pass about half of the worlds in the benchmark, a decidedly non-trivial achievement given the complexity of the task. 
These frontier models represent a substantial improvement over their predecessors, not just in reasoning over given data but in actively designing informative experiments. As shown in \autoref{fig:guided_vs_rng}, \texttt{claude-opus-4-7} and \texttt{gpt-5.5} outperform their own randomized-experiment baselines, with \texttt{claude-opus-4-7} in particular showing monotonic gains in pass rate and explanation score with increasing round budget. This widening gap between guided and random experimentation indicates that the benchmark rewards genuine long-horizon reasoning over an experimental history.

On the other hand, even the strongest models fail to solve the more difficult worlds, which are characterized by important latent structure (e.g. three particle species, dark matter, and extra dimensions).
As we show with several LLM-proposed explanations and discovered laws in \autoref{sec:appendix_examples}, failure is caused primarily by insufficient experimentation (see Figures \ref{fig:aha} and \ref{fig:fail} for an instructive example, where not probing enough measurement times to find the time-varying force law leads to a failure).  A more systematic per-run analysis of every failure by the two strongest agents, organised along seven capability axes, is given in \autoref{app:failure_analysis}.

We find a significant skill gap between proprietary and open-source models. Open-source models show little difference between guided and random experiments, suggesting they do not meaningfully exploit their experimental budget. The most performant open model, \texttt{Qwen-3.5}, solves only about $10\%$ of the worlds in five attempts and scores low on both trajectory MSE and explanation relative to the proprietary variants.

Our dual evaluation also reveals meaningful differences amongst the strongest proprietary models. \texttt{gpt-5.5} achieves the lowest trajectory MSEs usually without achieving the highest explanation scores. It also gains less explanation score through multiple experiment rounds compared to \texttt{claude-opus-4-7}, which we interpret as a tendency to lock in a candidate law early and refine its parameters rather than revise its conceptual picture, i.e. fitting the data well without necessarily understanding it.  The capability-gap decomposition in \autoref{app:failure_analysis} is consistent with this picture: \texttt{gpt-5.5} failures concentrate on considering the right kind of law in the first place, whereas \texttt{claude-opus-4-7} failures spread more across experimental design and self-monitoring.

\paragraph{Limitations and future extensions.} \textsc{DiscoverPhysics} is limited in  world generation and evaluation, both of which suggest future extensions that would increase its usefulness. 
The worlds, while non-canonical, are deliberately curated rather than genuinely novel: the underlying physics is unusual but designed by us, and the systems are small in scale, with few particles, no instrumental systematics, and no observational realism beyond a small amount of Gaussian noise. The set of worlds is also still modest in size. Although the simulator is built to accept arbitrary force laws, extending the suite requires human effort to design conceptually interesting worlds and write scoring rubrics for them.

Our evaluation methodology has limitations as well. The explanation score relies on a single LLM judge (\texttt{claude-opus-4-6}); ideally this would be averaged over multiple judge models to reduce any single-model bias. The fact that the judge ranks \texttt{gpt-5.5} above \texttt{claude-opus-4-7} on MSE and rates them comparably at low round budgets (\autoref{fig:guided_vs_rng}) is reassuring evidence against naive Claude-on-Claude favoritism, but does not fully resolve the concern. The pass/fail thresholds we use (10\% normalized MSE, 0.9 explanation score) are also somewhat arbitrary; we mitigate this by reporting mean explanation scores and normalized MSE alongside pass@k, so that aggregate trends do not hinge on a particular cutoff.

Finally, all models show pass@5 substantially larger than pass@1, which suggests that reinforcement learning could meaningfully improve scores on this task. We caution, however, that this is evidence RL can improve performance on \textsc{DiscoverPhysics} specifically, not that the resulting capabilities would generalize to scientific discovery more broadly. Assessing such transfer would require further work.

\section{Conclusion}

We present the benchmark \textsc{DiscoverPhysics} to test the abilities of agentic models to discover the physical laws governing simulated worlds, whose physics deviates considerably from standard textbook cases. The agent has to propose experiments over several rounds that are carried out with a black-box N-body simulator, with the goal of describing, in words and Python code, the physical principles the simulated worlds follow. We assess the model performance with a dual evaluation scheme, which comprises a conventional predictive accuracy metric as well as an explanation score from a LLM that judges the proposed physical explanation against a expert human description of the true underlying physical relations. This dual assessment reflects the goal of physics research to decode through experiments the latent principles of our world in ways that generalize beyond existing observations. To our knowledge, \textsc{DiscoverPhysics} is the strongest test of how well LLMs perform this ill-defined, iterative research process of hypothesis generation and refinement on the basis of self-proposed experiments. We find that the largest, most recent LLMs perform best and discover the correct explanation for about 
50\% of the worlds within five attempts. However, they employ different research strategies. While \texttt{gpt-5.5} finds excellent predictive accuracy quickly by adopting flexible governing equations, \texttt{claude-opus-4-7} is able to refine its conceptual guesses and therefore find more accurate explanations for the most unconventional worlds, demonstrating that the act of designing insightful experiments is critical for discovery in the physical sciences.

\begin{ack}
The authors thank Polymathic AI for funding and support. We would also like to acknowledge the support of the Simons Foundation and of Schmidt Sciences. This work was supported in part by the AI2050 program at Schmidt Sciences (Grant G-25-70028). AGW thanks NSF CAREER IIS-2145492 and DARPA AIQ HR0011259006 for support. MW thanks NYU CDS, and the Flatiron Intitute for hosting as a visiting researcher during part of this work. LMS was partially supported by a grant from Coefficient Giving and by Polymathic AI.
\end{ack}

\bibliography{refs}
\bibliographystyle{unsrtnat}  

\newpage
\appendix
\section{Example experimentation process}
\begin{figure}[h]
    \centering
    \includegraphics[width=0.98\linewidth]{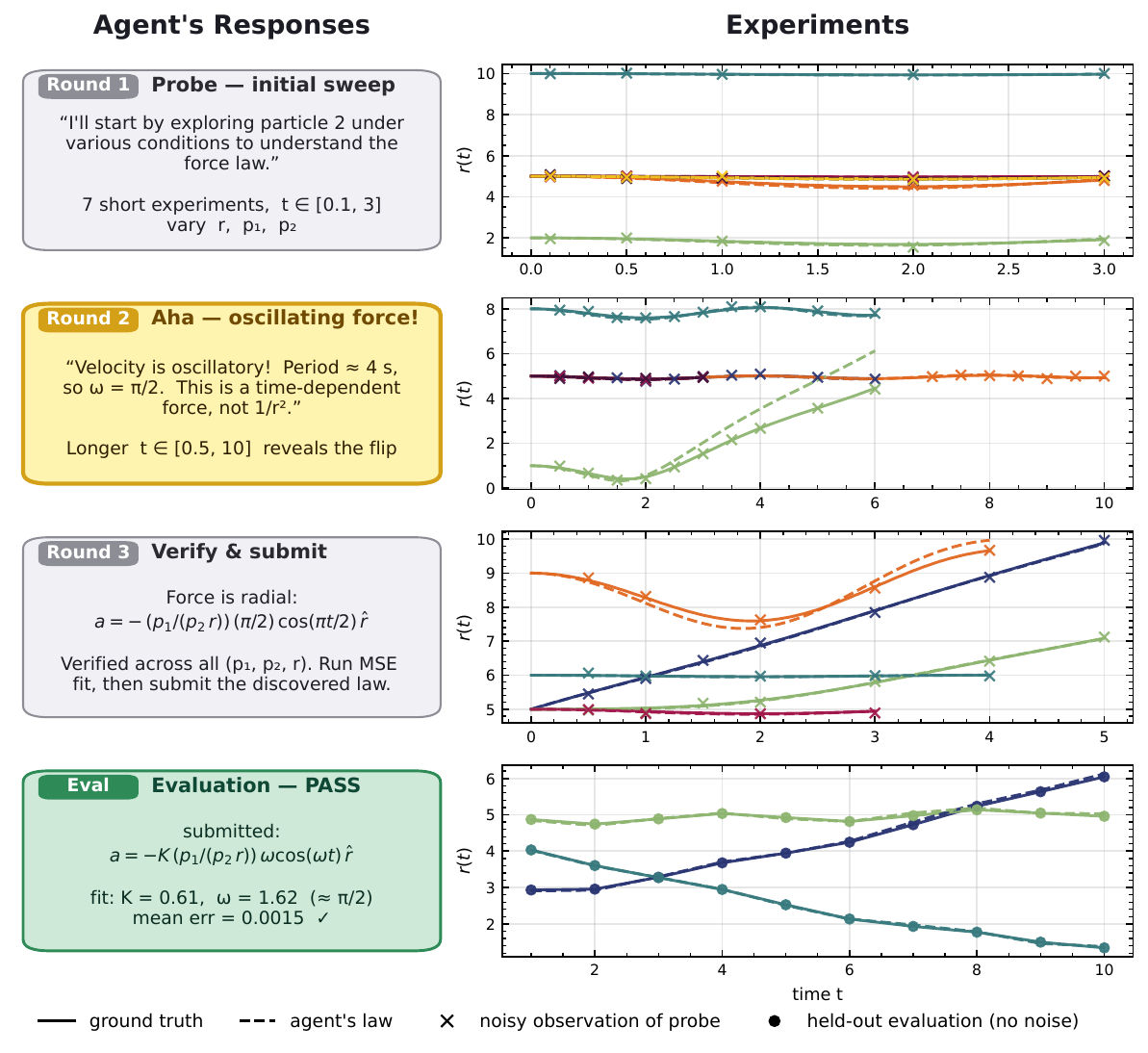}
    \caption{A summary of the experimentation process for an example run of Claude Opus 4.7 on the \texttt{oscillator} world. The model is ultimately successful in discovering the underlying force law due to long timescale experiments. We highlight the naive initial experiment, the ``aha moment'' of the model after it analyzes the naive experiment, the clever experimentation (choosing a timescale for its experiments that allows for the detection of the time-dependent force), and the finalization of its discovered law. The visualization of the model's experiments are shown on the right. $r(t)$ represents the position of the particles in the simulation, with the noisy observations of the model's proposed probe particles marked by an x, the ground-truth force law shown with solid lines, and the agent's proposed final law shown with dashed lines. The final evaluation is shown on held-out test particles the model never sees.}
    \label{fig:aha}
\end{figure}

\begin{figure}[h]
    \centering
    \includegraphics[width=0.98\linewidth]{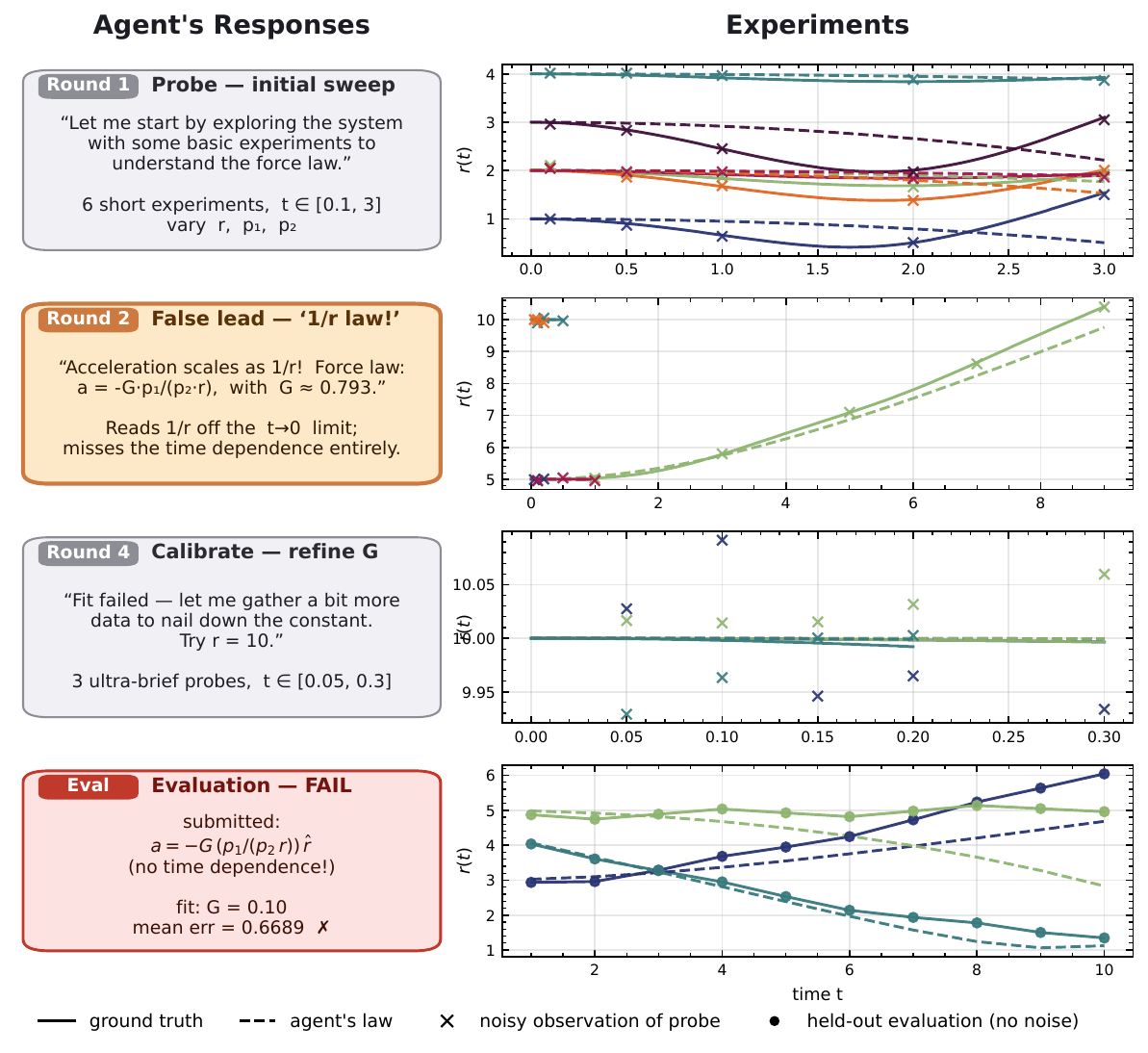}
    \caption{A summary of the experimentation process for an example run of Claude Opus 4.7 on the \texttt{oscillator} world. The model fails to discover the true underlying force law. We highlight the naive initial experiment, the false discovery of the model early on, and the smaller timescales in experimentation (missing the detection of the time-dependent force), and the finalization of its discovered law. The visualization of the model's experiments are shown on the right. $r(t)$ represents the position of the particles in the simulation, with the noisy observations of the model's proposed probe particles marked by an x, the ground-truth force law shown with solid lines, and the agent's proposed final law shown with dashed lines. The final evaluation is shown on held-out test particles the model never sees.}
    \label{fig:fail}
\end{figure}

In Figures \ref{fig:aha} and \ref{fig:fail} we show two examples of the experimentation process for Claude Opus 4.7 on the \texttt{oscillator} world, differing solely by the random seed we used for the agent. Both figures show a summary of a full run on this world with a maximum of 16 possible experimentation rounds.

In \autoref{fig:aha}, the agent decided to submit its final discovered law after four rounds. In round 1, the agent proposes an initial sweep of probe particle positions and times. It notices from the results of round 1 that the velocities of the probes are oscillating in time. It then decides to carefully probe the time-dependent force causing this oscillation in order to find the frequency and structure of the force in round 2. It submits much longer times to measure the probes at, and then in round 3, it analyzes these results and concludes that there must be explicit time-dependence in the force law, and proposes a fitting request to verify this. In round 4, it analyzes the results of the fit, is confident that it got the free parameters correct, and then submits its final answer.

In \autoref{fig:fail}, we contrast the successful experimentation in \autoref{fig:aha} with a failed attempt just by changing the random seed for sampling. Similarly, in round 1 the agent proposes an initial sweep of probe particle positions and times. 
It guesses a $\frac{1}{r}$ force law due to the behavior of the particles at short times. It then decides to submit experiments at very short timescales, where the time-dependent behavior is not detectable. In round 3, it submits an MSE fitting request, but there was an error and the fit did not go through. It chooses to continue to probe even smaller timescales in round 4 and continues to trust its ``discovered'' force law of $\frac{1}{r}$. In round 5, it is confident that it got the law correct, and then submits its final answer.

\clearpage

\section{Sample agent noise response}
\label{sec:appendix-noise}
A sample response from \texttt{claude-opus-4-7} on the Ether world at an observational noise level of $50\%$. The model clearly identifies noise as a potential issue, then determines the best experiments to run to minimize its effect on the results.
\begin{tcolorbox}[
    colback=gray!15!white,
    colframe=gray!95!white,
    sharp corners,
    boxrule=1pt,
    title=claude-opus-4-7 | noise level $50\%$ | Round 2 response,              
    fonttitle=\bfseries,          
    left=4pt, right=4pt,        
    top=4pt, bottom=4pt,      
]
\begin{quote}\itshape``Looking at the data more carefully now. The probe at $(20, 0)$ with mass $1$ --- at $t = 0.5$ it's near $(24.5, -4.1)$ and $(20.7, 0.8)$ in two trials. So at short times, there's significant noise ($\sigma$ maybe $\sim 3\text{--}4$?). At later times the positions vary wildly between identical experiments --- large noise.

Wait --- looking again. Two identical experiments give very different positions at $t = 20$. That suggests either:
\begin{enumerate}
    \item Huge observation noise ($\sigma \sim$ several units)
    \item The system itself is chaotic
\end{enumerate}

Let me focus on velocities, which may be less noisy. Actually for the probe at $(20, 0)$, $v = 0$, at $t = 0.5$ in run 1: $v \approx (-0.06, -1.16)$. Run 2: $v \approx (1.20, 0.01)$. Different!

That's chaotic behavior, but with $v = 0$ initial and almost no time, the velocity shouldn't change much from systematic forces in $0.5$s unless force is large. Means the noise must be on positions: $\sigma \approx 2\text{--}3$ units? And velocity reports are also noisy.

Let me check the ring orbiters. Particle 1 starts at $(5, 0)$ with $v = (0, 2.8209)$. For Keplerian orbit $GM = v^2 r = 39.77$. Let's check if this ring is in circular orbit in run 1 at $t = 20$: particle 1 should be back near start (period if Kepler: $T = 2\pi r / v = 2\pi \cdot 5 / 2.82 = 11.13$s, so at $t = 20$ should be about $1.8$ periods around, angle $= 20 \cdot v / r = 11.28$ rad $= 1.8$ rev).

Looking at reported position for particle 1 at $t = 20$ in run 1: $(1.28, 2.43)$. Should be at angle $11.28$ rad $\bmod 2\pi = 4.997$ rad $= 286^\circ$, so $\cos = 0.286$, $\sin = -0.958$, expected $(1.43, -4.79)$. Hmm, doesn't match closely. But position is on a circle $\sim r = 5$? $\sqrt{1.28^2 + 2.43^2} = 2.75$. Way too small.

Hmm. So either the orbit isn't circular gravity, or noise is enormous and position reports are very corrupted.

Let me think: the velocities for the ring orbiters are different too --- $v = 2.82$ is set up but at $t = 0$ isn't given. Let me check at $t = 0.5$: particle 1 reported velocity $(-1.59, 5.09)$. That's pretty different from initial $(0, 2.82)$. So actually large changes happen.

This is hard to disentangle with noise. Let me try: a single very heavy probe far away (so its motion is slow) with zero velocity, look at very short times, and use repeated runs to average.

Actually let me think differently. The noise $\sigma$ seems large --- maybe $\sim 3$ on positions? Position $(20, 0)$ goes to $(24.57, -4.13)$ in $0.5$s. If force were $\sim 1/r^2$ with $GM = 40$, accel $= 40/400 = 0.1$, so displacement in $0.5$s $= 0.0125$ --- utterly invisible against noise $\sigma \sim 3$.

The strategy must be: very long durations with large signal, and use many trials to average. Let me check if the ring (orbiters) are actually in a closed circular orbit, by averaging the radius across all $20$ orbiters at various times.''\end{quote}
\end{tcolorbox}

\section{Description of physics worlds} 
\label{sec:appdescr}

Worlds use one of four JSON configs. All of them are sent to the N-body simulator, where they are processed in the same way. The exact JSON keys for each config are defined in a prompt and substituted into the universal template via a \texttt{\{world\_instructions\}} placeholder.

\textbf{Two-particle ($p_1, p_2$).} Particle 1 is a fixed source at the origin; particle 2 is a mobile probe. The agent sets $p_1$ (which the simulator maps to the source coupling of particle 1), $p_2$ (the inertia of particle 2), $\mathbf r_2(0)$, $\mathbf v_2(0)$, and the measurement times. \texttt{gravity}, \texttt{yukawa}, \texttt{fractional}, \texttt{oscillator},
\texttt{extra\_dimensions}, and \texttt{coulomb\_easy} all use this set-up.

\textbf{Probe topologies.} \texttt{three\_species} (30 fixed background particles in three hidden species + 5 neutral probes; 35 total) and \texttt{dark\_matter} (20 visible + 10 hidden dark sources + 5 probes; 25 exposed to the agent, 35 total) accept the agent's 5 probe positions and velocities only. The probes have $s_i = 0$ (do not source the gravitational field) but $c_i \neq 0$ (do feel it).

\textbf{Anchor + ring + probes (with masses).} \texttt{ether} and \texttt{hubble} use 26 particles: a heavy central anchor (the only field source, $Q=50$), 20 ring orbiters with masses cycling through species $\{1,2,4\}$, and 5 probes whose positions, velocities, and masses (charges) the agent sets. These worlds are designed to test whether agents can disentangle a central field-mediated force from a body force supplied by space itself.

\textbf{Symmetric multi-body.} \texttt{species} (6 particles, two hidden source species), \texttt{circle} (1 center + 10 ring particles, all sourcing the field) place the agent in control of every particle's initial state (just radius and tangential velocity for the constrained \texttt{circle} world) with no fixed background.

\subsection{Simple gravity: \texttt{gravity}}
$1/r$ (logarithmic) attractive force in a 2D world.

\subsection{Screened potential: \texttt{yukawa}}
Yukawa potential: Short-range exponentially suppressed force, screening length $\lambda$.

\subsection{Fractional gravitation with 2 particles: \texttt{fractional}}
Fractional Laplacian $-(-\nabla^2)^\alpha$.

\subsection{Fractional gravitation with ring of particles: \texttt{circle}}
Fractional Laplacian $-(-\nabla^2)^\alpha$.

\subsection{Multi-species simulation: \texttt{three\_species}} 3 hidden classes (one repulsive) + 5 neutral probes.

\subsection{Dark matter (invisible mass): \texttt{dark\_matter}}
Hidden (from agent) dark halo.

\subsection{Luminiferous ether: \texttt{ether}}
Central law + a global preferred-direction drift.

\subsection{Hubble flow: \texttt{hubble}}
Central law  + radial Hubble flow $H(\mathbf{r})$ (position-dependent).

\subsection{Oscillator: \texttt{oscillator}}
Sinusoidally varying coupling ($G(t) = G_0\cos(\omega t + \varphi)$) that periodically reverses sign.

\subsection{Extra dimensions: \texttt{extra\_dimensions}}
$1/r$ to $1/r^2$ transition (2d+1 dimensions, reuses \texttt{gravity} world setup) with Kaluza-Klein image sums.

\subsection{Coulomb force: \texttt{coulomb\_easy}}
$1/r^2$ attractive force.

\section{Example rubric}
\label{sec:apprubric}
We provide an example of our scoring rubric for the worlds. This rubric is for the \texttt{yukawa} world. Scores range from 0 to 10, with 0 being the worst possible score and 10 the best.

\begin{tabular}{@{}p{0.08\textwidth}@{\hspace{1em}}p{0.84\textwidth}@{}}
\toprule
\textbf{Score} & \textbf{Criterion} \\
\midrule

\textbf{10} &
Identifies a static screened / Helmholtz operator,
\(\nabla^2 \phi - \frac{\phi}{\lambda^2} = \text{source}\),
or an equivalent Yukawa form; names or describes exponential suppression at long
range with screening length within roughly \(1\)--\(4\) (ground truth
\(\lambda = 2\)); correctly identifies \(p_1\) as source coupling and \(p_2\)
as inertia.
\\
\midrule

\textbf{7--9} &
Identifies screened / Yukawa-like behaviour qualitatively, such as short-range
attraction and long-range suppression, but misses a quantitative detail, for
example \(\lambda\) is badly off or not estimated, or muddles the roles of
\(p_1\) and \(p_2\).
\\
\midrule

\textbf{4--6} &
Recognises a static attractive field but treats it as a plain Laplacian or
generic power law and misses the screening structure, or asserts screening
without any characterisation of its scale.
\\
\midrule

\textbf{1--3} &
Uses the wrong operator family, such as a time-evolving or wave operator, or
gives no distance-dependent suppression despite the clear experimental
signature.
\\
\midrule

\textbf{0} &
Empty or irrelevant.
\\

\bottomrule
\end{tabular}

\section{Example outputs} 
\label{sec:appendix_examples}
We present a range of example outputs that are gathered after a round is finished, with different models and scores shown. The output consists of a discovered law in Python, an evaluation conducted by fitting the Python code with SciPy, a text explanation of the discovered physical law, and the LLM-judge's evaluation of the agent's text explanation (both in words and a numerical score out of 10). We show one example for each of the 11 public worlds.
\subsection{\texttt{gravity}}
\begin{lstlisting}[style=logstyle,caption={Final output after giving the science agent a budget of 16 rounds of experiments for a gravity world with Claude Opus 4.7 as the science agent.},label={lst:grav}]
============================================================
Discovered law:
============================================================
def discovered_law(pos1, pos2, p1, p2, velocity2, duration, **params):
    """
    2D logarithmic-potential interaction: particle 1 (held fixed at origin) sources
    a 2D Coulomb/gravitational field whose potential grows as ln(r), the natural
    Green's function of the 2D Laplacian. The probe particle 2 feels an attractive
    radial acceleration a = -(p1/(2 pi p2)) * r_vec / r^2.
    """
    import numpy as np
    from scipy.integrate import solve_ivp
    k = params.get("k", 1.0/(2.0*np.pi))
    x0, y0 = pos2[0], pos2[1]
    vx0, vy0 = velocity2[0], velocity2[1]
    def rhs(t, y):
        x, yy, vx, vy = y
        r2 = x*x + yy*yy
        if r2 < 1e-12:
            return [vx, vy, 0.0, 0.0]
        coef = -k * p1 / p2 / r2
        return [vx, vy, coef*x, coef*yy]
    sol = solve_ivp(rhs, (0.0, duration), [x0, y0, vx0, vy0],
                    method="RK45", rtol=1e-7, atol=1e-9)
    return [sol.y[0,-1], sol.y[1,-1]], [sol.y[2,-1], sol.y[3,-1]]

def fit_parameters():
    return {"k": {"init": 0.15915494309, "bounds": [0.05, 0.5]}}

============================================================
Evaluation:
============================================================
  [fit using 4/14 training trajectories (cap: 4 traj × 5 times)]
  Fitted parameters: k=0.1907
  Training-set loss: 0.007444 → 0.006899
  Case 1: mean_particle_mse = 0.0671
  Case 2: mean_particle_mse = 0.2032
  Case 3: mean_particle_mse = 0.0192

  Mean particle MSE: 0.0965
  Max  particle MSE: 1.0753
  Result: PASS

============================================================
Explanation evaluation:
============================================================
Agent explanation: Particle 1 sits at the origin and sources a 2D logarithmic potential φ(r) = (p1/(2π)) ln(r) — the Green's function of the Laplacian in two dimensions. Particle 2 couples to this field through its inertia p2, experiencing an attractive radial force F = -p1/(2π r) (*@$\hat{r}$@*) and hence acceleration a = -p1/(2π p2) · (*@$\hat{r}$@*)/r. This is the genuine 2D analogue of Newtonian gravity (force ∝ 1/r rather than 1/r²), and the prefactor 1/(2π) is the universal 2D Green's-function constant.

  Explanation score: 1.00  (raw 10.0/10)
  Judge reasoning: The student's description is essentially perfect. They correctly identify the static 2D Poisson/Laplacian equation, the logarithmic Green's function, the 1/r force law (not 1/r²), p1 as the source coupling strength, and p2 as the inertia of particle 2. The inclusion of the 1/(2π) prefactor is an accurate additional detail. All physical content matches the ground truth precisely.

<score>10</score>
\end{lstlisting}
\subsection{\texttt{yukawa}}
\begin{lstlisting}[style=logstyle,caption={Final output after giving the science agent a budget of 16 rounds of experiments for a yukawa world with Deepseek-R1-0528 as the science agent.},label={lst:yukawa}]
============================================================
Discovered law:
============================================================
from scipy.integrate import solve_ivp
import numpy as np

def discovered_law(pos1, pos2, p1, p2, velocity2, duration, **params):
    """
    Particle 2 experiences a central force from particle 1.
    The force magnitude scales with the ratio p1/p2 and
    decays as the inverse 1.5 power of distance.
    This creates stable orbits with period-distance
    relationships distinct from inverse-square gravity.
    """
    k = params.get("k", 0.118)
    
    def rhs(t, state):
        x, y, vx, vy = state
        r = np.sqrt(x**2 + y**2)
        r_vec = np.array([x, y])
        force_mag = k * (p1 / p2) / (r**1.5)
        ax = -force_mag * x / r
        ay = -force_mag * y / r
        return [vx, vy, ax, ay]
    
    y0 = [pos2[0], pos2[1], velocity2[0], velocity2[1]]
    sol = solve_ivp(rhs, (0, duration), y0, method='RK45', rtol=1e-4, atol=1e-4)
    
    final_pos = [sol.y[0][-1], sol.y[1][-1]]
    final_vel = [sol.y[2][-1], sol.y[3][-1]]
    return final_pos, final_vel

def fit_parameters():
    return {
        "k": {"init": 0.118, "bounds": [0.01, 1.0]}
    }

============================================================
Evaluation:
============================================================
  [fit using 4/25 training trajectories (cap: 4 traj × 5 times)]
  Fitted parameters: k=0.1561
  Training-set loss: 0.09799 → 0.09725
  Case 1: mean_particle_mse = 0.0427
  Case 2: mean_particle_mse = 0.1907
  Case 3: mean_particle_mse = 0.0121

  Mean particle MSE: 0.0818
  Max  particle MSE: 0.7869
  Result: FAIL

============================================================
Explanation evaluation:
============================================================
Agent explanation: The system features a central force law where particle 2 experiences acceleration proportional to p1/p2 and inversely proportional to distance raised to 1.5. The force is always attractive and radial from particle 1's fixed position. This fractional exponent creates orbital dynamics with period scaling as T ∝ (r^{1.25}) / (*@$\sqrt{p_1/p_2}$@*), distinct from inverse-square gravity.

  Explanation score: 0.30  (raw 3.0/10)
  Judge reasoning: The student describes a power-law force (1/r^1.5) that is "always attractive" with no mention of exponential suppression or screening. They miss the Helmholtz/Yukawa character entirely, treating it as a generic fractional power-law rather than a screened field. While they correctly identify p1 as source strength and p2 as inertial coupling, and recognize the force as static and attractive, the complete absence of screening/exponential suppression and the wrong operator family (fractional power-law instead of Helmholtz) place this solidly in the lower-middle range.

<score>3</score>
\end{lstlisting}

\subsection{\texttt{fractional}}

\begin{lstlisting}[style=logstyle,caption={Final output after giving the science agent a budget of 16 rounds of experiments for a fractional world with Qwen3.5-397B-A17B as the science agent.},label={lst:frac}]
============================================================
Discovered law:
============================================================
def discovered_law(pos1, pos2, p1, p2, velocity2, duration, **params):
    """Particle 2 experiences an attractive central force toward particle 1 (at origin).
    The acceleration magnitude scales as G * p1 / p2 / r^2, directed radially inward.
    This is an inverse-square law with coupling proportional to p1/p2 ratio.
    """
    import numpy as np
    from scipy.integrate import solve_ivp
    
    G = params.get("G", 0.16)
    
    def rhs(t, y):
        x, y_pos, vx, vy = y
        r = np.sqrt(x**2 + y_pos**2)
        if r < 1e-10:
            r = 1e-10
        factor = G * (p1 / p2) / (r**2)
        ax = -factor * x / r
        ay = -factor * y_pos / r
        return [vx, vy, ax, ay]
    
    y0 = [pos2[0], pos2[1], velocity2[0], velocity2[1]]
    sol = solve_ivp(rhs, (0.0, duration), y0, method="RK45", rtol=1e-6, atol=1e-6)
    final_state = sol.y[:, -1]
    final_pos2 = [final_state[0], final_state[1]]
    final_vel2 = [final_state[2], final_state[3]]
    return final_pos2, final_vel2

def fit_parameters():
    return {
        "G": {"init": 0.16, "bounds": [0.05, 0.5]},
    }

============================================================
Evaluation:
============================================================
  [fit using 4/38 training trajectories (cap: 4 traj × 5 times)]
  Fitted parameters: G=0.1695
  Training-set loss: 0.004411 → 0.004382
  Case 1: mean_particle_mse = 0.0006
  Case 2: mean_particle_mse = 0.0005
  Case 3: mean_particle_mse = 0.0003

  Mean particle MSE: 0.0005
  Max  particle MSE: 0.0021
  Result: PASS

============================================================
Explanation evaluation:
============================================================
Agent explanation: Particle 2 experiences an attractive central force directed toward particle 1 (fixed at the origin). The force follows an inverse-square law: the acceleration magnitude is G × (p1/p2) / r², where r is the distance from the origin. The scalar properties p1 and p2 act as coupling constants—p1 (source) enhances the attraction while p2 (probe) reduces the response, analogous to charge and mass respectively. The constant G ≈ 0.16 sets the overall force scale in this universe.

  Explanation score: 0.50  (raw 5.0/10)
  Judge reasoning: The student correctly identifies the system as static with an attractive central force, and correctly identifies p1 as the source coupling and p2 as reducing the response (inertia). However, the student describes the force as following an inverse-square law (1/r²), which corresponds to a standard 2D Laplacian (or 3D Coulomb-like) scenario, completely missing the fractional Laplacian operator and the enhanced long-range (slower decay) character that is the key distinguishing feature of this world. This places the answer squarely in the "recognises static attraction but assumes a standard Laplacian or ordinary power-law" band.

<score>5</score>
\end{lstlisting}

\subsection{\texttt{circle}}
\begin{lstlisting}[style=logstyle,caption={Final output after giving the science agent a budget of 16 rounds of experiments for a circle world with gpt-oss-120b as the science agent.},label={lst:circle}]
============================================================
Discovered law:
============================================================
def discovered_law(positions, velocities, duration, **params):
    """
    Particles experience a constant outward radial acceleration.
    The acceleration magnitude A is a free parameter (default 0.07).
    No other forces act; thus motion follows Newton's second law with
    a = A * (*@$\hat{r}$@*), where (*@$\hat{r}$@*) is the unit vector from the origin to the particle.
    """
    import numpy as np
    from scipy.integrate import solve_ivp

    # number of particles (should be 11)
    n = len(positions)

    # flatten initial state: [x1,y1,...,xn,yn, vx1,vy1,...,vxn,vyn]
    y0 = np.hstack([np.ravel(positions), np.ravel(velocities)])

    A = params.get("A", 0.07)  # outward acceleration magnitude

    def rhs(t, y):
        # unpack
        pos = y[: 2 * n].reshape((n, 2))
        vel = y[2 * n :].reshape((n, 2))

        # compute radial unit vectors, avoid division by zero
        r_norm = np.linalg.norm(pos, axis=1, keepdims=True)
        # for the central particle (index 0) keep acceleration zero
        with np.errstate(divide="ignore", invalid="ignore"):
            unit = np.where(r_norm > 0, pos / r_norm, 0.0)

        acc = A * unit  # constant outward acceleration
        dydt = np.concatenate([vel.ravel(), acc.ravel()])
        return dydt

    sol = solve_ivp(rhs, (0.0, duration), y0, method="RK45", rtol=1e-4, atol=1e-4)
    final_state = sol.y[:, -1]
    final_pos = final_state[: 2 * n].reshape((n, 2))
    return final_pos.tolist()

def fit_parameters():
    return {
        "A": {"init": 0.07, "bounds": [0.01, 0.2]}
    }

============================================================
Evaluation:
============================================================
  [fit using 4/8 training trajectories (cap: 4 traj × 5 times)]
  Fitted parameters: A=0.01
  Training-set loss: 141 → 44.19
  Case 1 (r=5.0, v_t=0.3): mean_particle_mse = 6.7159

  Mean particle MSE (all particles): 6.7159
  Max  particle MSE:                 24.9273
  Result: FAIL

============================================================
Explanation evaluation:
============================================================
Agent explanation: The hidden field is a uniform radial repulsion: every particle feels a constant acceleration directed away from the origin. The acceleration magnitude A is the sole free constant. With no torques, angular momentum is conserved, so particles follow outward spirals determined by their initial radial and tangential velocities. This simple central-force law reproduces the observed linear growth of speed and quadratic increase of radius.

  Explanation score: 0.10  (raw 1.0/10)
  Judge reasoning: The student describes a "uniform radial repulsion" with constant acceleration away from the origin, which is fundamentally wrong in multiple ways: the force is repulsive rather than attractive, there is no particle-particle interaction described (only a central force from the origin), no spatial operator (Laplacian, fractional Laplacian, or otherwise) is identified, and the static non-local fractional character is entirely missed. The description bears essentially no resemblance to the ground truth.

<score>1</score>
\end{lstlisting}
\subsection{\texttt{three\_species}} 
\begin{lstlisting}[style=logstyle,caption={Final output after giving the science agent a budget of 16 rounds of experiments for a three-species world with Qwen3-235B-A22B-Instruct-2507 as the science agent.},label={lst:3spec}]
============================================================
Discovered law:
============================================================
def discovered_law(positions, velocities, duration, **params):
    """
    The force on each probe is proportional to the 90-degree counterclockwise
    rotation of both its velocity and position vectors, indicating a synthetic
    magnetic-like field coupled with a rotational potential field.
    
    This results in trajectories that combine cyclotron motion with
    a background swirl, consistent with a fixed vorticity field.
    """
    import numpy as np
    from scipy.integrate import solve_ivp

    # Extract the 5 probe particles (last 5)
    probe_positions = np.array(positions[-5:])
    probe_velocities = np.array(velocities[-5:])

    # Set parameters
    alpha = params.get("alpha", 1.0)  # velocity coupling
    beta = params.get("beta", 1.0)    # position coupling

    # Initial state: [x1, y1, x2, y2, ..., vx1, vy1, ...]
    y0 = np.concatenate([probe_positions.flatten(), probe_velocities.flatten()])

    def rhs(t, y):
        pos = y[:10].reshape(5, 2)
        vel = y[10:].reshape(5, 2)
        acc = np.zeros_like(vel)

        # Rotate velocity and position 90° CCW: (a,b) -> (-b, a)
        acc += alpha * np.array([-vel[:, 1], vel[:, 0]]).T
        acc += beta * np.array([-pos[:, 1], pos[:, 0]]).T

        dydt = np.concatenate([vel.flatten(), acc.flatten()])
        return dydt

    # Solve ODE
    sol = solve_ivp(rhs, (0.0, duration), y0, method="RK45", rtol=1e-4, atol=1e-4, t_eval=[duration])
    final_state = sol.y[:, -1]
    final_positions = final_state[:10].reshape(5, 2)
    return final_positions.tolist()

def fit_parameters():
    return {
        "alpha": {"init": 1.0, "bounds": [0.1, 10.0]},
        "beta": {"init": 1.0, "bounds": [0.1, 10.0]}
    }

============================================================
Evaluation:
============================================================
  Case 1: ERROR -- operands could not be broadcast together with shapes (5,2) (35,2) 
  Case 2: ERROR -- operands could not be broadcast together with shapes (5,2) (35,2) 

  Mean particle MSE (all particles): inf
  Max  particle MSE:                 inf
  Result: FAIL

============================================================
Explanation evaluation:
============================================================
Agent explanation: The system is governed by a gyroscopic force law where each probe experiences an acceleration proportional to the 90-degree counterclockwise rotation of both its velocity and position vectors. This indicates a synthetic magnetic-like field coupled with a background rotational potential, resulting in spiral-like trajectories even from rest. The force does not originate from inter-particle interactions but from a fixed background field with intrinsic vorticity.

  Explanation score: 0.00  (raw 0.0/10)
  Judge reasoning: The student describes a gyroscopic/magnetic-like force law with velocity-dependent rotational forces, which is entirely wrong — the ground truth is a static Laplacian field with potential-based forces. The student mentions no inter-particle interactions, no species structure, no coupling strengths or signs, and no neutral probes. The description is fundamentally incorrect in every major aspect: wrong operator, wrong force law, wrong physical picture.

<score>0</score>
\end{lstlisting}

\subsection{\texttt{dark\_matter}}
\begin{lstlisting}[style=logstyle,caption={Final output after giving the science agent a budget of 16 rounds of experiments for a dark-matter world with GPT-5.5 as the science agent.},label={lst:dark}]
============================================================
Discovered law:
============================================================
def discovered_law(positions, velocities, duration, **params):
    """
    The background particles form a rigid population moving in an isotropic harmonic trap, so their trajectories are sinusoidal rotations/oscillations about the centre.
    The neutral probes do not contribute to the field, but they accelerate in the external central trap plus the superposed field generated by the moving background particles.
    Each background particle contributes an attractive softened power-law force to each probe, with the coupling, softening, and exponent left fit-adjustable.
    """
    import numpy as np
    from scipy.integrate import solve_ivp

    omega = params.get("omega", 0.255)
    k = params.get("k", 0.0846)
    G = params.get("G", 0.387)
    eps = params.get("eps", 0.21)
    beta = params.get("beta", 2.25)

    p0 = np.array(positions, dtype=float)
    v0 = np.array(velocities, dtype=float)
    n = p0.shape[0]
    nb = n - 5

    if duration <= 0:
        return p0.tolist()

    bg_p0 = p0[:nb]
    bg_v0 = v0[:nb]
    pr0 = p0[nb:]
    vr0 = v0[nb:]

    def bg_pos(t):
        c = np.cos(omega * t)
        s = np.sin(omega * t)
        return bg_p0 * c + (bg_v0 / omega) * s

    y0 = np.concatenate([pr0.ravel(), vr0.ravel()])

    def rhs(t, y):
        pr = y[:10].reshape((5, 2))
        vr = y[10:].reshape((5, 2))
        bp = bg_pos(t)
        d = bp[None, :, :] - pr[:, None, :]
        r2 = np.sum(d * d, axis=2) + eps * eps
        a = -k * pr + G * np.sum(d / (r2 ** (beta / 2.0))[:, :, None], axis=1)
        return np.concatenate([vr.ravel(), a.ravel()])

    sol = solve_ivp(rhs, (0.0, float(duration)), y0, method="RK45", rtol=1e-4, atol=1e-4)
    yf = sol.y[:, -1]
    prf = yf[:10].reshape((5, 2))
    final_positions = np.zeros_like(p0)
    final_positions[:nb] = bg_pos(float(duration))
    final_positions[nb:] = prf
    return final_positions.tolist()

def fit_parameters():
    return {
        "omega": {"init": 0.255, "bounds": [0.20, 0.32]},
        "k": {"init": 0.0846, "bounds": [-0.05, 0.25]},
        "G": {"init": 0.387, "bounds": [-5.0, 5.0]},
        "eps": {"init": 0.21, "bounds": [0.01, 3.0]},
        "beta": {"init": 2.25, "bounds": [0.8, 4.5]}
    }

============================================================
Evaluation:
============================================================
  Case 1: mean_probe_mse = 16.8422
  Case 2: mean_probe_mse = 26.8562

  Mean probe MSE (probes only): 21.8492
  Max  probe MSE:               147.8362
  Result: FAIL

============================================================
Explanation evaluation:
============================================================
Agent explanation: The background population moves as independent harmonic-oscillator particles about the centre, effectively rotating with a common angular frequency. The probes are passive test particles: they feel a central harmonic restoring acceleration and the sum of softened attractive power-law forces from the instantaneous positions of all background particles, while exerting no force back on anything. The main structural feature is therefore a time-dependent external field produced by a rigidly evolving background configuration plus a smooth central trap.

  Explanation score: 0.20  (raw 2.0/10)
  Judge reasoning: The student describes the system as harmonic oscillators with a central trap and softened attractive power-law forces, which is fundamentally different from the static 2D Laplacian field equation. The student fails to identify the Laplacian operator, does not recognize the existence of hidden/dark-matter sources (instead attributing motion to a "rigidly evolving background configuration plus a smooth central trap"), and while they correctly note probes are passive test particles, the overall mechanism is wrong. The student misses the key structural feature of hidden dark-matter particles causing anomalous accelerations.

<score>2</score>
\end{lstlisting}
\subsection{\texttt{ether}}
\begin{lstlisting}[style=logstyle,caption={Final output after giving the science agent a budget of 16 rounds of experiments for an ether world with Claude Sonnet 4.6 as the science agent.},label={lst:ether}]
============================================================
Discovered law:
============================================================
def discovered_law(positions, velocities, masses, duration, **params):
    """
    2D Newtonian gravity where force magnitude scales as G*m_i*m_j/r (not 1/r^2).
    Derived from 2D Gauss law: logarithmic potential U = -G*m_i*m_j*ln(r).
    G confirmed via short-time probe acceleration and ring circular orbit: G*M_anchor=7.96.
    """
    import numpy as np
    from scipy.integrate import solve_ivp

    G = params.get("G", 7.96e-15)
    pos0 = np.array(positions, dtype=float)
    vel0 = np.array(velocities, dtype=float)
    m = np.array(masses, dtype=float)
    N = len(pos0)
    y0 = np.concatenate([pos0.flatten(), vel0.flatten()])

    def rhs(t, y):
        p = y[:2*N].reshape(N, 2)
        v = y[2*N:].reshape(N, 2)
        dx = p[np.newaxis, :, :] - p[:, np.newaxis, :]
        r2 = np.sum(dx**2, axis=2)
        np.fill_diagonal(r2, 1.0)
        inv_r2 = 1.0 / r2
        np.fill_diagonal(inv_r2, 0.0)
        acc = G * np.einsum('j,ijd->id', m, dx * inv_r2[:, :, np.newaxis])
        return np.concatenate([v.flatten(), acc.flatten()])

    sol = solve_ivp(rhs, (0.0, duration), y0, method="RK45", rtol=1e-6, atol=1e-6)
    final = sol.y[:, -1]
    return final[:2*N].reshape(N, 2).tolist()

def fit_parameters():
    return {
        "G": {"init": 7.96e-15, "bounds": [1e-15, 1e-13]},
    }

============================================================
Evaluation:
============================================================
  [fit using 4/12 training trajectories (cap: 4 traj × 5 times)]
  [fit time budget 180s exceeded; using best-so-far params (loss=inf)]
  [fit hit 180s wall-clock budget; using best-so-far parameters]
  Fitted parameters: G=7.96e-15
  Training-set loss: inf → inf
  Case 1: mean_probe_mse = 35.7694
  Case 2: mean_probe_mse = 1.5856

  Mean probe MSE (probes only): 18.6775
  Max  probe MSE:               335.5333
  Result: FAIL

============================================================
Explanation evaluation:
============================================================
Agent explanation: In this 2D universe, gravity obeys a logarithmic potential law rather than the familiar 3D inverse-square law. The gravitational force between any two particles is F = G·m₁·m₂/r, directed along the line connecting them — this follows from applying Gauss's law in 2D where field lines spread over a circumference (2πr) rather than a spherical surface. The gravitational constant G ≈ 7.96×10⁻¹⁵ was confirmed both by measuring the initial acceleration of stationary probes at known distances from the massive anchor (a·r = G·M_anchor = 7.96 = constant) and by the circular orbit condition of the ring particles (v² = G·M_anchor = 2.82² ≈ 7.96). All 26 particles participate as full N-body gravitational sources, with the anchor (mass 1e15) dominating and the ring orbiters (masses cycling 1,2,4) contributing smaller perturbations.

  Explanation score: 0.40  (raw 4.0/10)
  Judge reasoning: The student correctly identifies the 2D Laplacian (logarithmic potential, 1/r force law) and the static nature of the field, and correctly identifies the anchor as the dominant central source with orbiter masses cycling through {1,2,4}. However, the student claims all 26 particles are "full N-body gravitational sources," whereas the ground truth specifies that only the anchor (index 0) has nonzero source coupling (50), with orbiters and probes being test particles (zero source coupling). Most critically, the student completely fails to identify the uniform northward drift acceleration (the "ether" body force F = α·m·ŷ), which is a key structural feature of this world. Missing the drift entirely places this squarely in the 4–6 band per the rubric.

<score>4</score>
\end{lstlisting}

\subsection{\texttt{hubble}}
\begin{lstlisting}[style=logstyle,caption={Final output after giving the science agent a budget of 16 rounds of experiments for a hubble world with Claude Opus 4.7 as the science agent.},label={lst:hub}]
============================================================
Discovered law:
============================================================
def discovered_law(positions, velocities, masses, duration, **params):
    """The anchor (particle 0) sources a static radial field with logarithmic
    short-range attraction and harmonic long-range repulsion: a = -A/r + B*r toward
    the origin. All other particles (orbiters and probes) are inert test particles
    that do not exert forces on each other; their inertial mass does not affect
    their acceleration in this field."""
    import numpy as np
    from scipy.integrate import solve_ivp
    A = params.get("A", 8.0)
    B = params.get("B", 0.05)
    pos = np.array(positions, dtype=float)
    vel = np.array(velocities, dtype=float)
    N = len(pos)
    anchor_pos = pos[0].copy()
    y0 = np.concatenate([pos.ravel(), vel.ravel()])
    def rhs(t, y):
        p = y[:2*N].reshape(N, 2)
        v = y[2*N:].reshape(N, 2)
        dv = np.zeros_like(p)
        for i in range(1, N):
            r_vec = p[i] - anchor_pos
            r = np.sqrt(r_vec[0]*r_vec[0] + r_vec[1]*r_vec[1])
            if r < 1e-8:
                continue
            a_rad = -A/r + B*r
            dv[i, 0] = a_rad * r_vec[0] / r
            dv[i, 1] = a_rad * r_vec[1] / r
        return np.concatenate([v.ravel(), dv.ravel()])
    sol = solve_ivp(rhs, (0.0, duration), y0, method="RK45", rtol=1e-7, atol=1e-9, max_step=0.5)
    final = sol.y[:, -1]
    final_pos = final[:2*N].reshape(N, 2)
    return final_pos.tolist()

def fit_parameters():
    return {
        "A": {"init": 8.0, "bounds": [0.5, 30.0]},
        "B": {"init": 0.05, "bounds": [0.001, 0.5]},
    }

============================================================
Evaluation:
============================================================
  [fit using 4/9 training trajectories (cap: 4 traj × 5 times)]
  [fit time budget 180s exceeded; using best-so-far params (loss=inf)]
  [fit hit 180s wall-clock budget; using best-so-far parameters]
  Fitted parameters: A=8, B=0.05
  Training-set loss: nan → nan
  Case 1: mean_probe_mse = 0.0184
  Case 2: mean_probe_mse = 0.0305

  Mean probe MSE (probes only): 0.0244
  Max  probe MSE:               0.1760
  Result: PASS

============================================================
Explanation evaluation:
============================================================
Agent explanation: The anchor at the origin sources a static, radial scalar field with potential V(r) = A·ln(r) − (B/2)·r², producing acceleration (*@$\vec{a}$@*) = (−A/r + B·r)·(*@$\hat{r}$@*) on every other particle. Short-range behavior is logarithmic attraction (∝ 1/r toward the anchor), while long-range behavior is harmonic repulsion (∝ r outward), with an unstable equilibrium at r = (*@$\sqrt{A/B}$@*). All non-anchor particles (background orbiters and probes) are passive test particles in this field — they do not source forces on one another, and their inertial mass cancels out so that acceleration depends only on position relative to the anchor (best-fit constants A ≈ 8, B ≈ 0.05).

  Explanation score: 1.00  (raw 10.0/10)
  Judge reasoning: The student correctly identifies a static central logarithmic (2D Laplacian) attraction sourced solely by the anchor, with all other particles as passive test particles — matching the ground truth. They also correctly identify the outward repulsive component as growing linearly with distance (∝ r, i.e., harmonic repulsion), which is equivalent to the Hubble-flow body-force a = H·r, and estimate B ≈ 0.05, matching H ≈ 0.05. They recognize the unstable equilibrium radius (critical radius) and the mass-independence of the acceleration. The estimate A ≈ 8 corresponds to Q/(2π) ≈ 8, giving Q ≈ 50, consistent with the ground truth. The only minor gap is not explicitly naming it a "Hubble flow" or identifying the specific mass cycling (1, 2, 4) of orbiters, but the physics is captured accurately and comprehensively.

<score>10</score>
\end{lstlisting}

\subsection{\texttt{oscillator}}
\begin{lstlisting}[style=logstyle,caption={Final output after giving the science agent a budget of 16 rounds of experiments for a oscillator world with GPT-5.5 as the science agent.},label={lst:osc}]
============================================================
Discovered law:
============================================================
def discovered_law(pos1, pos2, p1, p2, velocity2, duration, **params):
    """
    Particle 1 generates a central logarithmic field whose gradient falls as one over distance.
    The source strength is proportional to p1, while the probe inertial response is proportional to 1/p2.
    The coupling oscillates in time as cos(pi t / 2), causing alternating attraction and repulsion with period four.
    """
    import math
    from scipy.integrate import solve_ivp

    G = 0.8
    omega = math.pi / 2.0
    x1 = float(pos1[0])
    y1 = float(pos1[1])
    y0 = [float(pos2[0]), float(pos2[1]), float(velocity2[0]), float(velocity2[1])]

    def rhs(t, y):
        x, yy, vx, vy = y
        dx = x - x1
        dy = yy - y1
        r2 = dx * dx + dy * dy
        if r2 < 1e-12:
            return [vx, vy, 0.0, 0.0]
        fac = -G * float(p1) / float(p2) * math.cos(omega * t) / r2
        return [vx, vy, fac * dx, fac * dy]

    sol = solve_ivp(rhs, (0.0, float(duration)), y0, method="RK45", rtol=1e-7, atol=1e-9)
    yf = sol.y[:, -1]
    return [float(yf[0]), float(yf[1])], [float(yf[2]), float(yf[3])]

============================================================
Evaluation:
============================================================
  Case 1: mean_particle_mse = 0.0000
  Case 2: mean_particle_mse = 0.0000
  Case 3: mean_particle_mse = 0.0000

  Mean particle MSE: 0.0000
  Max  particle MSE: 0.0000
  Result: PASS

============================================================
Explanation evaluation:
============================================================
Agent explanation: Particle 1 creates a central logarithmic potential, so the force direction is radial and its magnitude scales as \(1/r\), equivalently acceleration is proportional to \(-\mathbf{r}/r^2\). The coupling strength is \(0.8\,p_1/p_2\), and the whole field is modulated by a temporal factor \(\cos(\pi t/2)\), giving a four-time-unit cycle of attraction and repulsion.

  Explanation score: 1.00  (raw 10.0/10)
  Judge reasoning: The student correctly identifies the 2D Laplacian/logarithmic potential with 1/r force law, the sinusoidal time modulation cos(πt/2) with period 4, the sign-flipping behavior (attraction and repulsion), and the p1/p2 coupling structure. The effective coupling coefficient is stated as 0.8 p1/p2, which matches G₀/(2π) ≈ 5/(2π) ≈ 0.796, demonstrating correct numerical identification. The student also correctly notes the phase φ = 0 implicitly. The only minor omission is not explicitly stating G₀ = 5 separately or elaborating on the absolute-time dependence breaking time-translation invariance, but all essential physics is captured.

<score>10</score>
\end{lstlisting}

\subsection{\texttt{extra\_dimensions}}
\begin{lstlisting}[style=logstyle,caption={Final output after giving the science agent a budget of 16 rounds of experiments for an extra-dimensions world with gpt-oss-120b as the science agent.},label={lst:extra}]
============================================================
Discovered law:
============================================================
def discovered_law(pos1, pos2, p1, p2, velocity2, duration, **params):
    """
    The probe particle feels an attractive inverse‑square central force whose
    strength scales with the ratio of the probe and source scalar properties,
    plus a linear viscous drag opposing its motion.
    """
    import numpy as np
    from scipy.integrate import solve_ivp

    # fitted parameters (default values allow reasonable predictions without fitting)
    K = params.get("K", 0.81)          # central force prefactor
    gamma = params.get("gamma", 0.09)  # linear drag coefficient

    def rhs(t, y):
        x, y_pos, vx, vy = y
        r = np.hypot(x, y_pos) + 1e-12
        # central inverse‑square attraction, scaled by p2/p1
        a_c = -K * (p2 / p1) * np.array([x, y_pos]) / (r**3)
        # linear drag
        a_drag = -gamma * np.array([vx, vy])
        return np.concatenate((np.array([vx, vy]), a_c + a_drag))

    y0 = np.array([pos2[0], pos2[1], velocity2[0], velocity2[1]], dtype=float)
    sol = solve_ivp(rhs, (0.0, duration), y0, method="RK45", rtol=1e-5, atol=1e-5)
    xf, yf, vxf, vyf = sol.y[:, -1]
    return [xf, yf], [vxf, vyf]

def fit_parameters():
    return {
        "K": {"init": 0.81, "bounds": [0.1, 5.0]},
        "gamma": {"init": 0.09, "bounds": [0.0, 0.5]},
    }

============================================================
Evaluation:
============================================================
  [fit using 4/30 training trajectories (cap: 4 traj × 5 times)]
  Fitted parameters: K=1.048, gamma=0.01065
  Training-set loss: 3.855 → 3.592
  Case 1: mean_particle_mse = 3.9007
  Case 2: mean_particle_mse = 1.5727
  Case 3: mean_particle_mse = 6.5475

  Mean particle MSE: 4.0070
  Max  particle MSE: 23.8647
  Result: FAIL

============================================================
Explanation evaluation:
============================================================
Agent explanation: Particle 2 moves under a central attractive field emanating from particle 1. The field obeys an inverse‑square law, with a strength proportional to the ratio of the probe’s scalar property p2 to the source’s property p1. Its motion is additionally damped by a linear drag proportional to its velocity.

  Explanation score: 0.20  (raw 2.0/10)
  Judge reasoning: The student describes only a single inverse-square (1/r²) force law with no mention of any crossover, compactification, extra dimension, or deviation at different distance scales. They also incorrectly introduce a linear drag term not present in the ground truth and mischaracterize the coupling structure (claiming force is proportional to p2/p1 rather than p1 with p2 as inertia). This fits the rubric's 1–3 band: a single canonical law without acknowledging failure at other distance ranges.

<score>2</score>
\end{lstlisting}

\subsection{\texttt{coulomb\_easy}}
\begin{lstlisting}[style=logstyle,caption={Final output after giving the science agent a budget of 16 rounds of experiments for a coulomb world with Claude Haiku 4.5 as the science agent.},label={lst:coulomb}]
============================================================
Discovered law:
============================================================
def discovered_law(pos1, pos2, p1, p2, velocity2, duration, **params):
    """
    The particle 2 experiences a central attractive force towards particle 1 proportional to p1 and inversely proportional to the square of the distance.
    The force magnitude is F = k * p1 / r^2, directed radially inward, with effective unit mass giving acceleration a = F.
    The property p2 does not influence the dynamics.
    """
    import numpy as np
    from scipy.integrate import solve_ivp
    
    k = params.get("k", 0.04)
    
    def rhs(t, y):
        x, y_pos, vx, vy = y
        r = np.sqrt(x**2 + y_pos**2)
        if r < 1e-8:
            r = 1e-8
        a_mag = k * p1 / (r**2)
        ax = -a_mag * x / r
        ay = -a_mag * y_pos / r
        return [vx, vy, ax, ay]
    
    y0 = [pos2[0], pos2[1], velocity2[0], velocity2[1]]
    sol = solve_ivp(rhs, (0.0, duration), y0, method="RK45", rtol=1e-4, atol=1e-4)
    
    final_state = sol.y[:, -1]
    return [final_state[0], final_state[1]], [final_state[2], final_state[3]]

def fit_parameters():
    return {
        "k": {"init": 0.04, "bounds": [0.001, 0.5]},
    }

============================================================
Evaluation:
============================================================
  [fit skipped: no training trajectories available]
  Case 1: mean_particle_mse = 3.7517
  Case 2: mean_particle_mse = 3.2569
  Case 3: mean_particle_mse = 3.6573

  Mean particle MSE: 3.5553
  Max  particle MSE: 15.0256
  Result: FAIL

============================================================
Explanation evaluation:
============================================================
Agent explanation: Particle 2 experiences a Coulomb-like central attractive force towards the fixed particle 1 at the origin. The force magnitude scales as F = k·p1/r², where r is the distance between particles, k ≈ 0.04 is a coupling constant, and p1 is the source particle's property. The acceleration equals the force (unit effective mass), and the direction is always radially inward towards particle 1. The property p2 does not affect the particle's dynamics.

  Explanation score: 0.50  (raw 5.0/10)
  Judge reasoning: The student correctly identifies a static central attractive 1/r² force with particle 1 fixed at the origin and unit effective mass for particle 2. However, the student explicitly states that p2 does not affect the dynamics, which contradicts the ground truth where the force is proportional to the product p1·p2. The student writes F = k·p1/r² instead of F = k·p1·p2/r², missing the role of p2 as a charge. The coupling constant estimate of k ≈ 0.04 is far from 1 (more than a factor of 2 off). The student gets the qualitative behavior and the 1/r² scaling correct but fails on the coupling structure regarding p2.

<score>5</score>
\end{lstlisting}

\section{Failure mode analysis of frontier agents}
\label{app:failure_analysis}

We next study the specific failure modes and capability gaps that current frontier models exhibit on this benchmark.  For the two strongest agents, \texttt{claude-opus-4-7} and \texttt{gpt-5.5}, we performed a per-run analysis of every failure on the eleven paper worlds, aiming to go beyond simplly checking whether the MSE results in pass/fail and identify why each agent did not recover the ground-truth law.

\subsection{Method}

Each benchmark run consists of up to sixteen rounds of experimentation followed by a submitted \texttt{discovered\_law} Python function.  We analyse every run in which the MSE threshold was not met: 30 of 55 runs for \texttt{claude-opus-4-7} (45\% pass rate) and 20 of 55 for \texttt{gpt-5.5} (64\% pass rate).

The analysis is an LLM-mediated \emph{summarise-then-classify} pipeline.  A first pass compresses the multi-thousand-line run transcript into a short structured per-run summary; a second pass reads that summary and assigns the run to a single primary capability gap from a predefined list.  Both passes use \texttt{claude-opus-4-7}.  Using the same model as one of the subjects of the analysis is a caveat; we mitigate it by releasing every per-run output so the categorisations can be audited or regenerated with a different analyser.

\paragraph{Stage 1: structured summary.}  For each failed run the analyser reads the full transcript, including the agent's reasoning at each round, the experiments it proposed, the data it received, the candidate laws it submitted for parameter fitting, its final submitted law, and the judge's ground-truth explanation.  It emits a short JSON record containing (a) a one-sentence summary of the submitted law, (b) a one-sentence description of the key mistake, and (c) one to six specific failure-mode descriptions grounded in the trace.  This record is the input to Stage~2.

\paragraph{Stage 2: capability attribution.}  In a separate pass the analyser reads only the Stage-1 JSON record for each run (it is permitted to fall back to the raw transcript if the summary is ambiguous) and assigns the run to a single primary capability from the seven-item list used in Table~\ref{tab:capability} and a single primary layer of the reasoning pipeline from the five-item list defined in the caption of Table~\ref{tab:layer}.  These two lists are a separate taxonomy from the Stage-1 scaffold; they sit at a higher level of abstraction (underlying cognitive capability rather than surface mistake), and were also hand-designed by us in advance.  Splitting the pipeline into two passes rather than classifying capabilities directly from the raw transcript is a practical choice: transcripts are long (often 10,000+ lines), while the Stage-1 summary is short enough that Stage~2 can be re-run against a different capability taxonomy without re-ingesting the transcripts.  Assigning a single primary category forces a pick in runs where two capabilities are co-implicated; the resulting counts should be read as a summary of where each failure most prominently lived, not as a claim of exclusivity.

\subsection{Capability gaps}
\label{app:capability_gaps}

\begin{figure}[h]
    \centering
    \includegraphics[width=\linewidth]{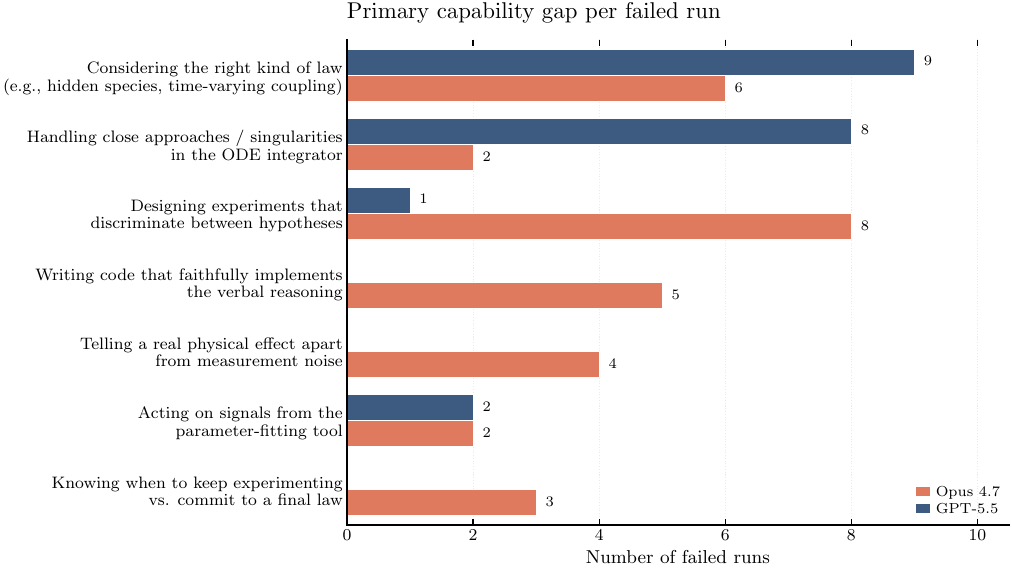}
    \caption{Visualisation of \autoref{tab:capability}.  Despite similar overall pass rates on the eleven paper worlds (45\% for Opus 4.7, 64\% for GPT-5.5), the two models appear to fail for qualitatively different reasons.  GPT-5.5 failures concentrate on considering the right kind of law and on robust numerical integration; Opus 4.7 failures spread across experimental design, writing code that matches its own verbal reasoning, telling signal apart from noise, and knowing when to stop experimenting.}
    \label{fig:capability_profile}
\end{figure}

We organise failures along seven capabilities that a successful discovery agent must combine.  The number of runs for which each capability was judged to be the primary bottleneck is shown in \autoref{tab:capability} and visualised in \autoref{fig:capability_profile}.

\begin{table}[h]
    \centering
    \caption{Primary capability gap for each of the 50 failed runs on the eleven paper worlds.  Each run is assigned to exactly one capability, namely the one whose absence could most plausibly change the outcome.  Opus 4.7 and GPT-5.5 appear to fail for quite different reasons despite comparable overall pass rates.}
    \label{tab:capability}
    \small
    \renewcommand{\arraystretch}{1.3}
    \begin{tabular}{p{0.62\linewidth} r r}
        \toprule
        \textbf{\shortstack[l]{Capability\\\small (what the agent failed to do)}} & \textbf{\shortstack{Opus 4.7\\(\# failed runs)}} & \textbf{\shortstack{GPT-5.5\\(\# failed runs)}} \\
        \midrule
        Consider the right kind of law & 6  & 9  \\
        \arrayrulecolor{black!30}\cmidrule(lr){1-3}
        Handle close approaches and singularities in the ODE integrator & 2  & 8  \\
        \cmidrule(lr){1-3}
        Design experiments that discriminate between competing hypotheses, rather than only measuring what the leading hypothesis predicts & 8  & 1  \\
        \cmidrule(lr){1-3}
        Write code that faithfully implements the verbal reasoning & 5  & 0  \\
        \cmidrule(lr){1-3}
        Tell a real physical effect apart from measurement noise, rather than dismissing a contradicting signal as noise & 4  & 0  \\
        \cmidrule(lr){1-3}
        Act on signals from the parameter-fitting tool & 2  & 2  \\
        \cmidrule(lr){1-3}
        Know when to keep experimenting versus commit to a final law & 3  & 0  \\
        \arrayrulecolor{black}\bottomrule
    \end{tabular}
\end{table}

A secondary classification records which layer of the reasoning pipeline was judged to be the locus of the failure.  These are enumerated in \autoref{tab:layer} and defined in the caption.

\begin{table}[h]
    \centering
    \caption{Reasoning-pipeline layer at which each failure was judged to occur.  The five layers are defined as follows.  \textit{Prior knowledge}: the agent never considered the right class of hypothesis.  \textit{Perception}: the relevant signal in the data was not extracted.  \textit{Reasoning}: the signal was extracted but the wrong inference was drawn from it.  \textit{Execution}: the reasoning was correct but the submitted code did not implement it.  \textit{Self-monitoring}: the agent generated or received a contradiction (a diagnostic from the fitter, a residual that did not match the model, a derivation that pointed elsewhere) and did not act on it.  Cell entries are the number of failed runs assigned to each layer.  Both models seem to accumulate execution-layer failures in similar numbers; Opus 4.7 appears relatively weaker at reasoning and self-monitoring, while GPT-5.5 appears relatively more prior-knowledge bound.}
    \label{tab:layer}
    \small
    \renewcommand{\arraystretch}{1.3}
    \begin{tabular}{p{0.62\linewidth} r r}
        \toprule
        \textbf{\shortstack[l]{Layer\\\small (where in the reasoning pipeline the failure occurred)}} & \textbf{\shortstack{Opus 4.7\\(\# failed runs)}} & \textbf{\shortstack{GPT-5.5\\(\# failed runs)}} \\
        \midrule
        Execution (right idea, wrong code)                                & 8  & 8  \\
        \arrayrulecolor{black!30}\cmidrule(lr){1-3}
        Reasoning (saw the data, drew the wrong inference)                & 10 & 5  \\
        \cmidrule(lr){1-3}
        Self-monitoring (generated a contradiction and did not act on it) & 8  & 2  \\
        \cmidrule(lr){1-3}
        Prior knowledge (never considered the right class of hypothesis)  & 3  & 5  \\
        \cmidrule(lr){1-3}
        Perception (relevant signal never extracted from the data)        & 1  & 0  \\
        \arrayrulecolor{black}\bottomrule
    \end{tabular}
\end{table}

At comparable pass rates, the two models seem to fail for qualitatively different reasons.  Opus 4.7's failures appear spread across \emph{experimental design}, \emph{writing code that matches the verbal reasoning}, \emph{telling signal apart from noise}, and \emph{knowing when to stop experimenting}: its prior seems adequate, and the discipline of the scientific process tends to be the weaker element.  GPT-5.5's failures appear to concentrate on \emph{considering the right kind of law} and \emph{robust numerical integration}: latent structure is often missed in the first place, and the implementation can be too rough to survive edge cases even when the form is right.

\subsection{Illustrative examples per capability}
\label{app:worked_examples}

We illustrate each capability with one or two short examples from the paper worlds.  These are not exhaustive; they are meant to ground the abstract category in a concrete observation.

\paragraph{Considering the right kind of law.}  This is the most frequent bottleneck in our sample and is implicated in all four worlds where both models fail universally.  On \texttt{three\_species}, which has hidden per-particle couplings $\{+1, +3, -2\}$, all ten failures submit a single uniform coupling; Opus 4.7 even derives a linear model of the observed $y$-asymmetry on one seed before dismissing it.  On \texttt{dark\_matter}, no run hypothesises the ten hidden sources: Opus 4.7 inflates the visible coupling to match the far-field, while GPT-5.5 replaces the point-source picture with a harmonic trap or prescribed background orbits.

\paragraph{Handling close approaches and singularities in the ODE integrator.}  This mode disproportionately affects GPT-5.5.  On \texttt{coulomb\_easy}, four of five GPT-5.5 runs identify the correct $p_1 p_2 / r^2$ force but fail on a held-out head-on radial passage through $r=0$; one submission uses no softening and slings a particle to $x = -252$, another uses a Plummer $\varepsilon = 0.02$ that does not match the evaluator's regularisation.

\paragraph{Designing experiments that discriminate between competing hypotheses.}  Opus 4.7 appears bottlenecked here more often than GPT-5.5.  On \texttt{gravity}, one seed concludes that the acceleration is independent of $p_2$ because no experiment isolated $p_2$ at fixed $(p_1, r)$, and drops the $1/p_2$ inertial factor from the submitted law.  On \texttt{three\_species}, both models sample the far-field monopole but never release a probe close to a single isolated source, which would have measured per-source coupling directly.

\paragraph{Writing code that faithfully implements the verbal reasoning.}  This mode seems almost unique to Opus 4.7 in our sample (five failures versus none for GPT-5.5): the agent describes the correct physics in prose but submits code that does not match the description.  On \texttt{yukawa}, one Opus 4.7 seed writes \texttt{r = sqrt(x**2 + y**2)}, hard-coding the source at the origin and breaking translation covariance on held-out cases; another has the $K_1(\alpha r)$ form right but hard-codes the amplitude as $\alpha / (2\pi)$ with $\alpha$ the only free parameter, so the external refit warps $\alpha$ from $0.5$ to $0.275$ to absorb the missing constant.

\paragraph{Telling a real physical effect apart from measurement noise.}  Also Opus 4.7-specific in our sample (four failures versus none for GPT-5.5).  On \texttt{dark\_matter} seed 3, Opus 4.7 measures a 6.7-fold near-field acceleration excess over the visibles-only model and attributes it to noise plus chaotic divergence of close encounters.  On \texttt{three\_species}, two seeds derive per-particle coupling structure from the data and discard it as overfitting.

\paragraph{Acting on signals from the parameter-fitting tool.}  A well-calibrated agent might treat plateaued losses or refits that barely move the loss as evidence that the candidate law is misspecified.  On \texttt{three\_species}, four consecutive GPT-5.5 fits plateau at loss $\approx 13$ without prompting a pivot to a different hypothesis class.  On \texttt{yukawa} seed 2, the external refit is forced to warp $\alpha$ by almost a factor of two to absorb a missing amplitude constant, but the submitted law is finalised without inspecting the refit.

\paragraph{Knowing when to keep experimenting versus commit to a final law.}  Opus 4.7 appears more prone to early submission than GPT-5.5 in our sample.  On \texttt{oscillator} seed 0, Opus 4.7 submits after two of sixteen rounds with a static central potential.  On \texttt{fractional} seed 3, it submits at round five while its fitted $k = 0.060$ is 2.6-fold off the analytic value $1/(2\pi) = 0.159$ that the agent itself computed.

\end{document}